# Enhancing the Parameterization of Reservoir Properties for Data Assimilation Using Deep VAE-GAN

Marcio A. Sampaio[a1], Paulo H. Ranazzi[a2], Martin J. Blunt[b3]

[a]Departamento de Engenharia de Minas e de Petróleo, Escola Politécnica, Universidade de São Paulo, 05508-030, SP, Brasil.

[b]Department of Earth Science and Engineering, Imperial College London, South Kensington, London, UK

[1]Corresponding author. E-mail address: marciosampaio@usp.br (Marcio A. Sampaio). ORCID: 0000-0003-1125-7218

[2]Contributing author. E-mail address: ranazzi@usp.br (Paulo H. Ranazzi). ORCID: 0000-0002-4515-4797

[3]Contributing author. E-mail address: m.blunt@imperial.ac.uk (Martin J. Blunt). ORCID: 0000-0002-8725-0250

## ARTICLE INFO

## Authorship contribution statement

Author 1: Conceptualization, Methodology, Data Curation, Formal analysis, Investigation, Software, Validation, Visualization, Writing. Author 2: Data curation, Datasets creation, Investigation, Validation, Writing. Author 3: Conceptualization, Data curation, Investigation, Validation, Writing, Visualization, Supervision.

ABSTRACT

Currently, the methods called Iterative Ensemble Smoothers, especially the method called Ensemble Smoother with Multiple Data Assimilation (ESMDA) can be considered state-of-the-art for history matching in petroleum reservoir simulation. However, this approach has two important limitations: the use of an ensemble with finite size to represent the distributions and the Gaussian assumption in parameter and data uncertainties. This latter is particularly important because many reservoir properties have non-Gaussian distributions. Parameterization involves mapping non-Gaussian parameters to a Gaussian field before the update and then mapping them back to the original domain to forward the ensemble through the reservoir simulator. A promising approach to perform parameterization is through deep learning models. Recent studies have shown that Generative Adversarial Networks (GAN) performed poorly concerning data assimilation, but generated more geologically plausible realizations of the reservoir, while the Variational Autoencoder (VAE) performed better than the GAN in data assimilation, but generated less geologically realistic models. This work is innovative in combining the strengths of both to implement a deep learning model called Variational Autoencoder Generative Adversarial Network (VAE-GAN) integrated with ESMDA. The methodology was applied in two case studies, one case being categorical and the other with continuous values of permeability. Our findings demonstrate that by applying the VAE-GAN model we can obtain high quality reservoir descriptions (just like GANs) and a good history matching on the production curves (just like VAEs) simultaneously.

## 1. INTRODUCTION

Ensemble methods represent the state-of-the-art for performing data assimilation in reservoir simulation models. In these methods, an ensemble of models is used to represent uncertain parameters and predictions. During the data assimilation process, the ensemble is adjusted to honor a set of available measurements. A comprehensive review of ensemble-based methods and their applications in history matching can be found in Aanonsen et al. (2009) and Oliver and Chen (2011). The Ensemble Smoother with Multiple Data Assimilation (ESMDA), introduced by Emerick and Reynolds (2013), is among the most widely used ensemble-based methods. The main disadvantages of ensemble-based methods are their poor performance in highly nonlinear models, spurious correlations from limited ensemble size, and reliance on Gaussian assumptions, which degrades performance when reservoir prior parameters exhibit non-Gaussian distributions, a common situation in large-scale reservoirs.

Many works have applied parameterization techniques in ensemble-based methods. Liu and Oliver (2005) parameterized geological facies using a Truncated Pluri-Gaussian (TPG) method. Since then, TPG and its variants have been investigated in several studies, such as Agbalaka and Oliver (2008), Zhao et al. (2008), Armstrong et al. (2011), Sebacher et al. (2013), and Sebacher et al. (2017). Other well-known parameterization methods are the level-set (Chang et al., 2010; Moreno and Aanonsen, 2011; Lorentzen et al., 2012; Ping and Zhang, 2014), normal-score (Zhou et al., 2011; Li et al., 2018), Discrete Cosine Transform (DCT) (Jafarpour and McLaughlin, 2008), and based on extensions of principal component analysis (PCA) (Sarma et al., 2008; Vo and Durlofsky, 2014; Chen et al., 2016; Emerick, 2017), among others.

Among the different parameterization approaches, deep learning (DL) techniques have recently gained significant attention in the community. Since the first known application by Canchumuni et al. (2017), the number of DL-based parameterization has grown rapidly. For parameterization in ensemble-based methods, the first DL applications were made by applying autoencoders, such as Canchumuni et al. (2017), where an autoencoder was tested to generate a continuous representation of a binary facies model for history matching with an ensemble smoother. Despite the good results obtained, the authors noticed that when the reservoir model became more complex the autoencoder was no longer able to represent facies with acceptable accuracy. Later, Canchumuni et al. (2019a), presented a comprehensive review and applications of deep belief networks in the parameterization problem, showing promising results. In Canchumuni et al. (2019b) results were reported for a Convolutional Variational Autoencoder (CVAE), where good data mismatch quality and good quality reservoir descriptions were presented for simple channeled models, but with lower quality for more complex 3D cases. Tadjer et al. (2021) combined machine learning techniques in the framework of data assimilation: t-Distributed Stochastic Neighbor Embedding (t-SNE) and Gaussian Process Latent Variable Model (GPLVM), with the addition of K-means clustering to define an effective ensemble to perform data assimilation. Razak et al. (2021) introduced another promising methodology, called Latent-Space Data Assimilation (LSDA), where both model parameters and data were compressed by an autoencoder. The assimilation was conducted in the latent space using a surrogate model to save computational time. However, it can also introduce nonlinearities that can degrade the assimilation performance. It is important to mention that most of the previously mentioned methods were applied in low-dimensional facies models, usually channeled reservoirs.

In the parameterization problem, the goal is to train the network to be able to generate samples of the data distributed approximately according to the same distribution of the training data. Among the deep generative models, generative adversarial networks (GANs) have also gained attention in the field of data assimilation. The first work that applied GANs of geological facies was carried out by Chan and Elsheikh (2017), with an extended version presented later (Chan and Elsheikh, 2019a). This work applied a Wasserstein GAN (Arjovsky et al., 2017) to the parameterization problem, with further improvements in conditional generation to honor the hard data (Chan and Elsheikh, 2019b). Zhang et al. (2021) proposed a different method to include conditional data in the generation of realizations, considering a U-Net as a generating network, where the inputs are now the latent space and a field containing the conditioning data. Canchumuni et al. (2021) analyzed different architectures of GANs, including transfer learning techniques, in the quality of assimilation of facies data and production data, showing promising results for their application. Several examples of GAN applications can be found in the literature (Mosser et al., 2018, 2019; Feng et al., 2022; Razak and Jafarpour, 2022; Zhang et al., 2022). Bao et al. (2022) compared the performance of GANs and Variational Autoencoders (VAEs) in the data assimilation problem, where the GAN performed poorly concerning data assimilation, but generated more realistic facies. The VAE performed better than the GAN in data assimilation, but generated unrealistic reservoir models. Finally, Ranazzi (2023) created a classifier neural network trained on synthetic reservoir descriptions to extract the necessary features for training a GAN with L1 regularization in the context of limited dataset size. By analyzing some generated cases, it was possible to observe the effect of entanglement that often occurs in GAN applications, although it generated good quality reservoir models. Results using ESMDA showed that some wells did not have good matches. The conjecture for these bad results for history matching is that the latent space generated by GAN was discontinuous and non-monotonic. By employing ESMDA, which calculates the update based on covariances, that are a linear measure of dependence, it fails to adequately adjust using this latent space.

To overcome the difficulties usually encountered when using GANs, such as mode collapse, training instability, entanglement, and the need for data augmentation, this research will use a trained VAE, alone or in conjunction with a GAN, so that the mapping between the latent space and the final reservoir/realization domain is less “entangled” (in terms of continuity and monotonicity), allowing the full functioning of the ESMDA in history matching. This is the main challenge to be overcome the main bottleneck in the integration of ESMDA with parameterization via deep learning models. In this way, the main targets of this work are generating geologically realistic reservoir descriptions from the GAN together with good matching (with lower errors) of the production profiles, improving on work where both targets could not be achieved simultaneously (Bao et al., 2022).

In the context of computational fluid dynamic (CFD) simulations, Cheng et al. (2020) proposed a hybrid deep adversarial autoencoder (VAE-GAN) for predicting parameterized nonlinear fluid flows using a flow past a cylinder and water column collapse as test cases. The results showed that VAE-GAN successfully captured spatio-temporal flow features, decreasing the computational time. Other works have applied VAE-GAN in different research areas, such as for synthesizing images such as faces of a specific person or objects in a category (Bao et al., 2017), to generate a variety of plausible two-dimensional magnetic topological structures data (Park et al., 2023), and for synthetic data generation in smart homes (Razghandi et al., 2022). All these applications showed advantages of the hybrid model in relation to the use of only one model (GAN or VAE).

In this work, we explore the advantages and disadvantages of the VAE-GAN model in parameterization for data assimilation with ensemble smoothers. The main contribution of this research is the improvement the generation of image facies with geologic realism in the same time with better data assimilation, trough the VAE-GAN hybrid model, combining the GAN's ability to generate high-quality images with the VAE network's capacity to produce higher-quality latent space for data assimilation.

## 2. METHODOLOGY

In this section, we present the methodology developed in this work. It consists by two main parts. The first part consists in modeling and training three deep learning models to compare the advantages and disadvantages of each one. The second part consists in make the integration with data assimilation, using the Ensemble Smoother with Multiple Data Assimilation (ESMDA) to perform history matching. Images or realizations in this study are a 2D map of permeability on a Cartesian grid.

## 2.1. Deep Learning Models

In this first part, we describe briefly three deep learning models used in this work: Generative Adversarial Network (GAN), Variational Autoencoder (VAE) and Variational Autoencoder Generative Adversarial Network (VAE-GAN).

### *2.1.1. Generative Adversarial Network (GAN)*

We utilize the Deep Convolutional GAN (DCGAN) (Radford et al., 2015), which optimizes the standard adversarial min-max objective (Goodfellow et al., 2014) using convolutional layers suited for image-based geological realizations. Information about the mathematical foundations of GANs can be found in Hong et al. (2020). During generator training, the discriminator weights are frozen, and the generator is penalized for failing to trick the discriminator, and only the generator weights are updated using the discriminator loss function. The structure of GAN is shown in Fig. 1.

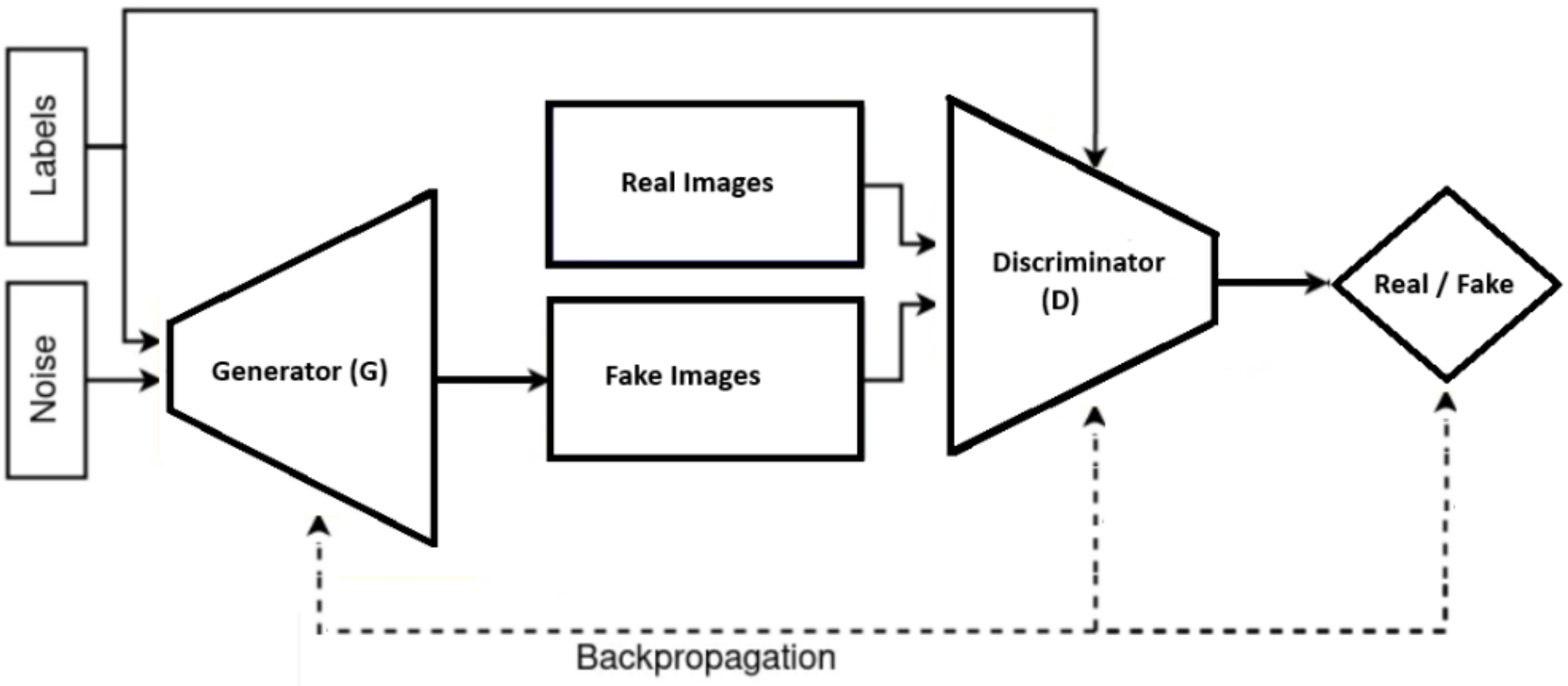

Fig. 1: A typical structure of a GAN.

In this way, each network has its own objective, where the convergence of this two-player game will be the Nash equilibrium. The problem can be formulated as the following min-max objective:

$$\min_{\mathcal{G}} \max_{\mathcal{D}} \mathbb{E}_{x \sim p_d}\left[\log\left(\mathcal{D}(x)\right)\right] + \mathbb{E}_{\hat{x} \sim p_g}\left[\log\left(1 - \mathcal{D}(\hat{x})\right)\right] \tag{1}$$

In this work, we employed the Deep Convolutional Generative Adversarial Network (DCGAN), an advanced version of the GAN that uses convolutional layers instead of fully connected layers for both the generator and discriminator. GANs can be used as parametrical models focused on learning how to generate samples from complex probability distributions (Canchumuni et al., 2021).

#### ***2.1.2. Variational Autoencoder (VAE)***

The deep learning model called Variational Autoencoder (VAE) was introduced by Kingma and Welling (2013) and is a type of generative model. A VAE has the same structural components as a traditional autoencoder: an encoder and a decoder. The encoder is a neural network responsible for mapping input data to a latent space. Unlike traditional autoencoders that produce a fixed point in the latent space, the encoder in a VAE outputs parameters of a probability distribution:

$$z \sim \mathcal{N}(\mu, \sigma^2) \tag{2}$$

where $z$ is the latent vector sampled from a Gaussian distribution ($\mathcal{N}$), using μ (mean) and $\sigma^2$ (variance) as parameters.

Since sampling $z$ directly is non-differentiable, VAE needs to use the reparameterization trick to perform the backpropagation:

$$z = \mu + \sigma \,.\, \mathcal{E} \,, \;\; \mathcal{E} \sim \mathcal{N}(0, 1) \tag{3}$$

This is particularly important for our problem, because we can force the latent space ($z$) to follow a normal Gaussian distribution $\mathcal{N}(0, 1)$, a main assumption of ESMDA in the data assimilation. This allows the VAE to model data uncertainty and variability effectively. Another neural network called a decoder is used to reconstruct the original data from the latent space representation. Given a sample from the latent space distribution, the decoder aims to generate an output that closely resembles the original input data. This process allows the VAE to create new data instances by sampling from the learned distribution. The latent space is a lower-dimensional, continuous space where the input data is encoded. The input data $(x)$ is fed into the encoder, which outputs the parameters of the latent space distribution $q(z \mid x)$. Latent variables $(z)$ are sampled from the distribution $q(z \mid x)$ using techniques including the reparameterization trick. The sampled $(z)$ is passed through the decoder to produce the reconstructed data $(\hat{x})$ from the distribution $q(\hat{x} \mid z)$, which should be like the original input $(x)$. Fig. 2 shows the typical structure of VAE.

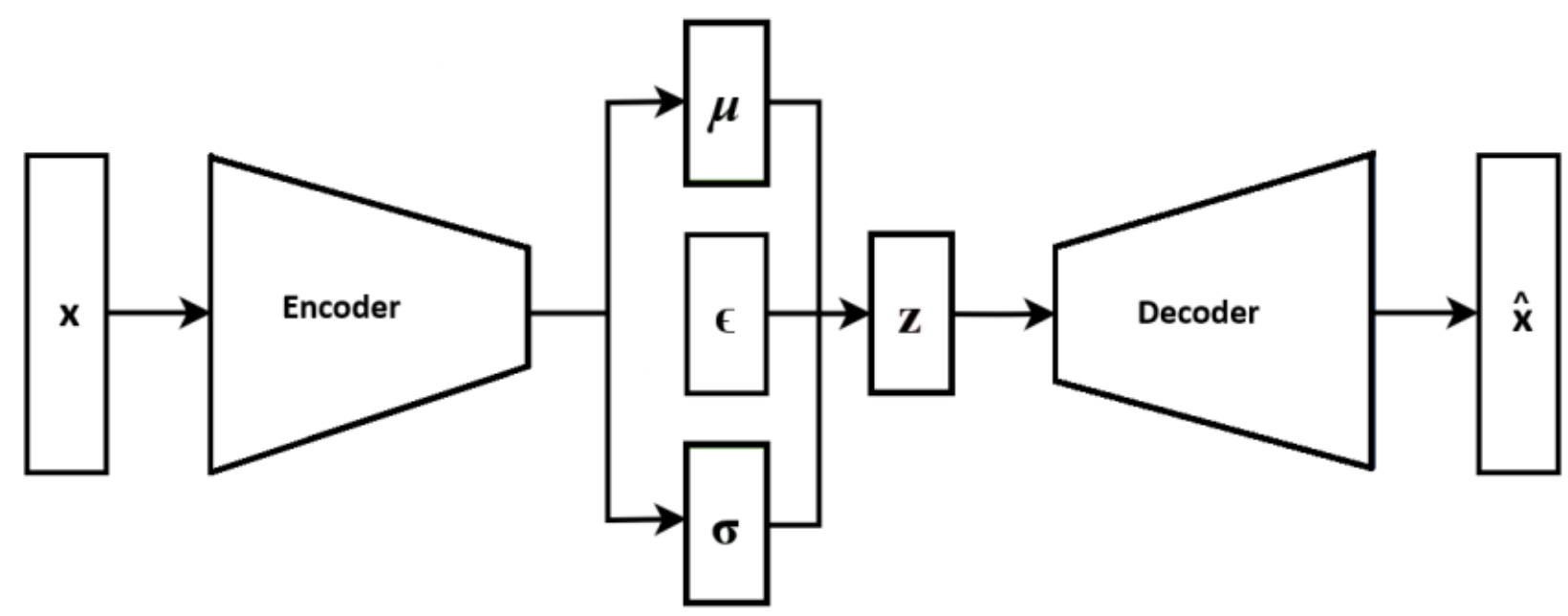

Fig. 2: A typical structure of VAE.

The VAE optimizes a loss function called the Evidence Lower Bound (ELBO), a combination of reconstruction loss and Kullback-Leibler (KL) divergence (Kullback and Leibler, 1951). The reconstruction loss measures how well the decoder reconstructs the input, using a metric, usually the Mean Square Error (MSE). And KL divergence regularizes the latent space by encouraging the learned distribution to be close to a standard normal distribution. The ELBO formulation is:

$$\mathcal{L} = \mathbb{E}_{z \sim q(z|x)}[\log p(x|z)] - KL(q(z|x) \parallel p(z)) \quad (4)$$

where $q(z|x)$ is the encoder's distribution, $p(z)$ is the prior (standard Gaussian) and $p(x|z)$ is the decoder's likelihood. The term KL can be multiplied by $\beta$ to increase or decrease the weight of this term.

In this work, we employed the Deep Convolutional Variational Autoencoder (DCVAE), a Variational Autoencoder (VAE) that uses convolutional layers in both the encoder and decoder networks to learn hierarchical, low-dimensional representations of high-dimensional data (in this case the reservoir description).

### *2.1.3. Variational Autoencoder Generative Adversarial Network (VAE-GAN)*

The Variational Autoencoder Generative Adversarial Network (VAE-GAN) combines the advantages of VAE and GAN to improve the quality of image generation while maintaining good learning of latent representations. The encoder of VAE learns a latent representation of the image and imposes a probabilistic distribution on this latent space. The decoder of VAE reconstructs images from the latent representation. The discriminator of the GAN distinguishes real images from images generated by the decoder. The VAE is responsible for ensuring that the latent space is well structured and regularized. GAN improves the quality of generated images by forcing the decoder to generate more realistic images. The discriminator is trained to distinguish real images from generated images. The VAE and GAN losses are combined to balance reconstruction and realism. Fig. 3 shows the VAE-GAN structure. Like GAN and VAE, in VAE-GAN we also employed convolutional layers to generate high-quality images.

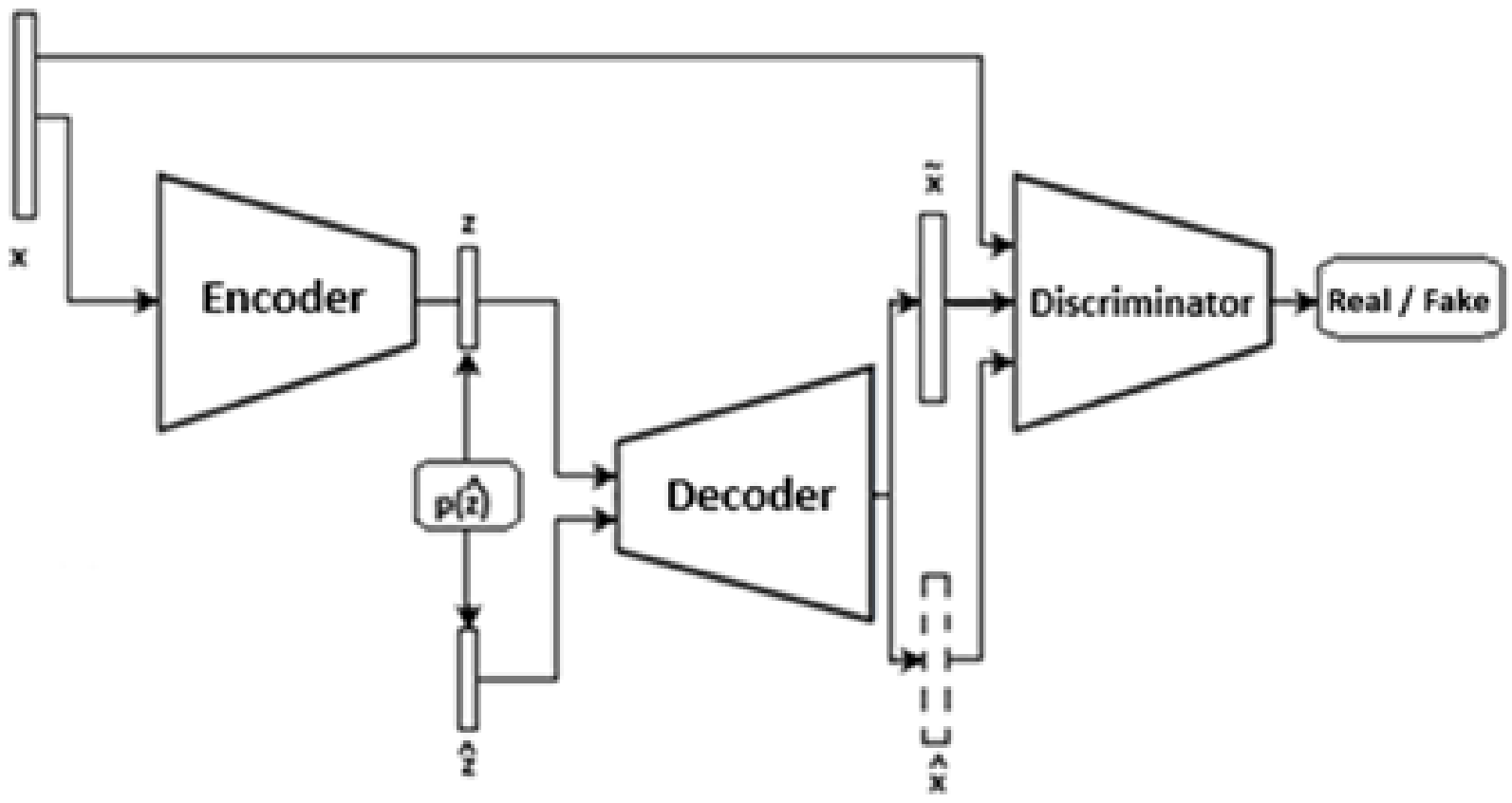

Fig. 3: A typical structure of VAE-GAN model.

## 2.2. Ensemble Smoother with Multiple Data Assimilation (ESMDA)

In this section, we will define the forward model and the inverse problem in history matching. The perfect forward model, where the model errors are neglected, can be defined as:

$$\mathbf{d} = g(\mathbf{m}) \quad (5)$$

where the forward operator g (in this case the reservoir simulator) relates the model parameters vector $\mathbf{m}_j \in \Re^{N_m}$ with the predicted data vector $\mathbf{d}_j \in \Re^{N_d}$, and, $N_m$ and $N_d$ are the number of model parameters and measurements, respectively. The history matching inverse problem is to estimate the model parameters that best represent the behavior of the available measurements of **d, hereafter defined as** $\mathbf{d}_{\text{obs}}$. We also have some sets of noisy measurements $\mathbf{d}_{\text{obs}}$ of the true data value $\mathbf{d}_{\text{true}}$:

$$\mathbf{d}_{\text{obs}} = \mathbf{d}_{\text{true}} + \epsilon \quad (6)$$

where the error $\epsilon$ is usually drawn from $\epsilon \sim \mathcal{N}(0, \mathbf{C}_{\text{D}})$, with $\mathbf{C}_{\text{D}} \in \Re^{N_d \times N_d}$ representing the covariance matrix of measurement errors.

Emerick and Reynolds (2013) defined that ES is equivalent to a full Gauss-Newton update step. Therefore, the ESMDA assimilates all the measurements multiple times, using an inflated covariance of measurement errors, relating one iteration with a smaller Gauss-Newton update step. In the ESMDA analysis step, each member *j* of model parameters vector ensemble is updated using the following update equation:

$$\mathbf{m}_j^{i+1} = \mathbf{m}_j^i + \mathbf{C}_{MD}^i \left(\mathbf{C}_{DD}^i + \alpha_i \mathbf{C}_{\text{D}}\right)^{-1} \left(\mathbf{d}_{obs,j}^i - \mathbf{d}_j^i\right) \quad (7)$$

for $j = (1, \ldots, N_e)$, where $N_e$ denotes the number of ensemble members (ensemble size), $\mathbf{C}_{MD}^i \in \Re^{N_m \times N_d}$ is the cross-covariance between the model parameters and predicted data, $\mathbf{C}_{DD}^i \in \Re^{N_d \times N_d}$ is the auto-covariance of the predicted data, $\alpha_i$ is the inflation factor that damps the iteration *i*. The forward step is where each ensemble member $\mathbf{d}_j^i$ is estimated using the forward operator $\mathbf{d}_j^i = g(\mathbf{m}_j^i)$ for $j = (1, \ldots, N_e)$.

### 2.3. Quality Metrics

In this section, we introduce the metrics used to assess the data assimilation and training results.

#### *2.3.1. Normalized data-mismatch objective function*

For each member $j$, the normalized data-mismatch is the difference between simulated and observed data, scaled by the measurements error covariance matrix:

$$\mathrm{O}_{\mathrm{N_d},\mathrm{j}} = \frac{1}{\mathrm{N_d}}\left(\mathbf{d}_j - \mathbf{d}_{\mathrm{obs}}\right)^{\mathrm{T}} \mathbf{C}_{\mathrm{D}}^{-1}\left(\mathbf{d}_j - \mathbf{d}_{\mathrm{obs}}\right) \tag{8}$$

and its average:

$$\overline{\mathrm{O}_{\mathrm{N_d}}} = \frac{1}{N_e}\sum_{\mathrm{j}=1}^{N_e} \mathrm{O}_{\mathrm{N_d},\mathrm{j}} \tag{9}$$

#### *2.3.2. Balanced Accuracy*

Balanced accuracy is a performance metric used in classification tasks, especially when dealing with imbalanced datasets. Balanced accuracy generalizes naturally to multi-class problems by averaging the recall (true positive rate) for each class. For $C$ classes, balanced accuracy is computed as:

$$\text{Balanced Accuracy} = \frac{1}{\mathrm{C}}\sum_{\mathrm{i}=1}^{\mathrm{C}} \frac{\text{True positives}_{\mathrm{i}}}{\text{True positives}_{\mathrm{i}} + \text{False negatives}_{\mathrm{i}}}, \text{for class i} \tag{10}$$

#### *2.3.3. Root Mean Square Error (RMSE)*

An ensemble of root mean square error (RMSE) for the ensemble $\mathbf{M} = \{\mathbf{m}_j\}_{j=1}^{\mathrm{N}_e}$ , with $M$ is the number of parameters and $N_e$ is the number of members of ensemble, with respect to the true model $\mathbf{m}_{\mathrm{true}}$ is:

$$\text{RMSE}\left(\mathbf{M}, \mathbf{m}_{\mathrm{true}}\right) = \left\{\frac{\left\|\mathbf{m}_j - \mathbf{m}_{\mathrm{true}}\right\|_2}{\sqrt{M}}\right\}_{j=1}^{N_e} \tag{11}$$

#### *2.3.4. Spread*

The spread is defined as the average RMSE between each ensemble member and the ensemble mean:

$$\text{Spread}(\mathbf{M}) = \frac{1}{N_e}\sum_{j=1}^{N_e} \frac{\left\|\mathbf{m}_j - \overline{\mathbf{m}}\right\|_2}{\sqrt{M}} \tag{12}$$

### *2.3.5. Fréchet Inception Distance (FID) and Fréchet Reservoir Distance (FRD)*

Since its introduction by Heusel et al. (2018), the Fréchet Inception Distance (FID) has become a standard two-sample test for evaluating the quality of generative models. The FID is estimated by computing the Fréchet distance between the distributions $r$ and $g$, obtained from the Inception network coding layer:

$$\text{FID}(r, g) = \left\|\mu_r - \mu_g\right\|_2^2 + Tr\left(\mathbf{C}_r + \mathbf{C}_g - 2\left(\mathbf{C}_r\mathbf{C}_g\right)^{\frac{1}{2}}\right) \tag{13}$$

where $\mu$ and $\mathbf{C}$ denote the mean and covariance of a given set of samples, while $Tr(\cdot)$ calculates the trace of a matrix. For further information about Inception network and GAN metrics, reader is directed to Borji (2018). Although widely used in computer vision, computing the FID metric to assess the quality of generated reservoir realizations poses challenges, because the nature of the ImageNet dataset is very different from the geological images used in this work. We replaced the Inception model with a Reservoir Classifier (RC) Network for computing the Fréchet Distance. Hereafter, this modified metric is referred to as the Fréchet Reservoir Distance (FRD). For more information on the Reservoir Classifier (RC) and Fréchet Reservoir Distance (FRD) used in this work, the reader can find all the details in Ranazzi (2023).

## 2.4. Integration Between Deep Learning Models and ESMDA

In this section, we present the integration between generative deep learning models and ESMDA. Thus, we use generative models to learn data representation and ESMDA to update the latent vector based on observational data. This was accomplished through two main steps:

Step 1: Training deep learning models to learn the distribution of the dataset. After training, the weights of the generator (GAN) and decoders (VAE and VAE-GAN) are saved for use in the next step;

Step 2: Employment of ESMDA to update the latent representations based on observed data. The prior latent vectors ($\mathbf{z}_o$) are the result of the encoders for a set of prior realizations generated with geostatistics. The ensemble of latent vectors is used in the decoder (or generator in the GAN case) to generate an ensemble of realizations which goes in the simulator to compute an ensemble of predicted data. The ESMDA is used do update the vector **z** and the process continues until the number of data assimilation iterations is reached. The complete workflow for integrating deep learning models with ESMDA is shown in Fig. 4 in terms only of latent vector for simplicity.

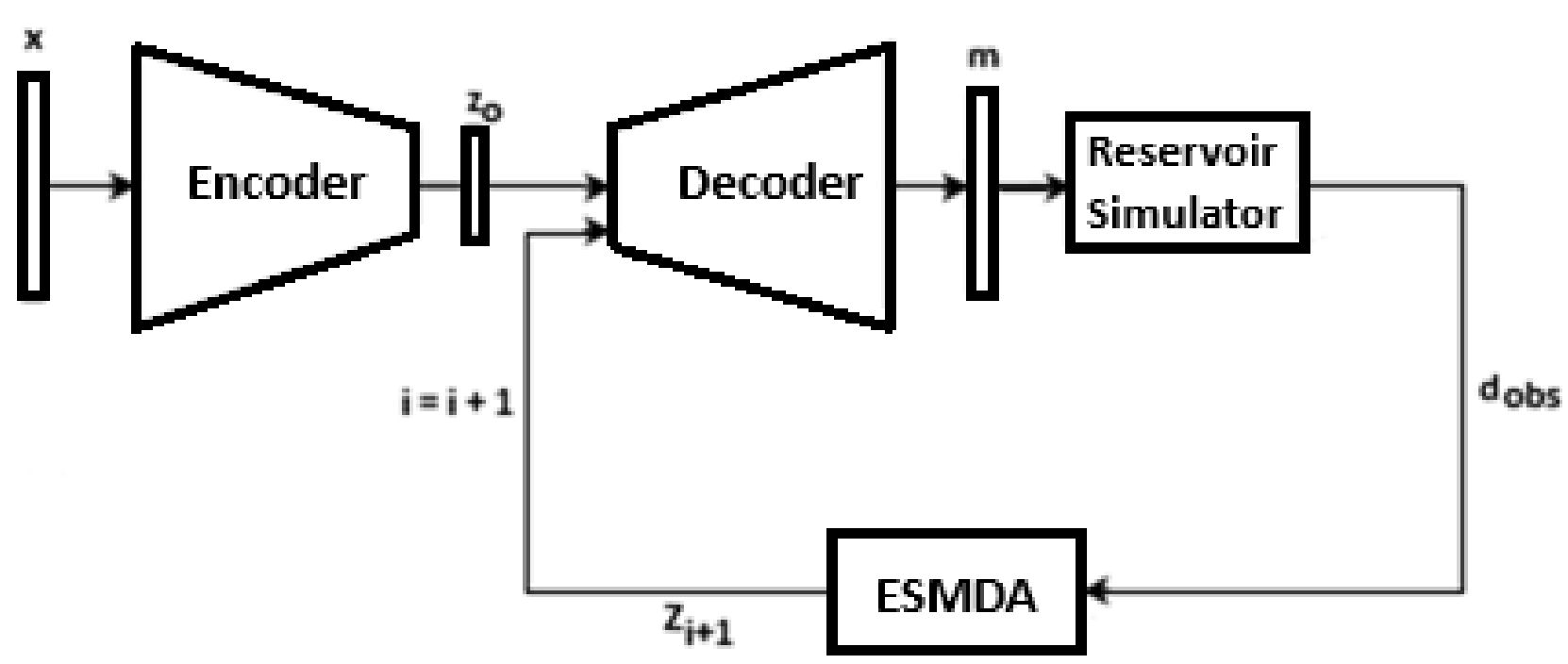

Fig. 4: The complete workflow of integration between generative models and ESMDA. In the DCGAN case, the representation of decoder is the generator network.

## 3. CASE STUDIES

We evaluate the methodology by considering two distinct synthetic datasets: one case consisting of categorical variables (integer values) representing facies, and the second one considering continuous variables based on the carbonate reservoir benchmark. The datasets were normalized to a range of -1 to 1, which is necessary for using the tanh activation function in the output layer of the generator network of GAN (Radford et al., 2015) or decoder in VAE and VAE-GAN models. All prior realizations, in each dataset, were generated using the same training image of the reference model.

### 3.1. Categorical training dataset

The realizations of the first dataset were built using the open-source Stanford Geostatistical Modeling Software – SGeMS (Remy et al., 2009), by applying the MPS algorithm Single Normal Simulation Equation – SNESIM (Strébelle, 2000, 2002). Using the "Stanford V" three-facies training image from Remy et al. (2009, chap. 8), we have created 80 000 categorical realizations, with a rotation angle of 45º in relation to the training image. In the images, shown here normalized for training, they have original permeability values equal to 100, 1000 and 9000 mD, for the colors purple, green and yellow, respectively. Fig. 5 shows some unconditional realizations obtained by SNESIM algorithm. Henceforth, this dataset referred to as the 'categorical training dataset'.

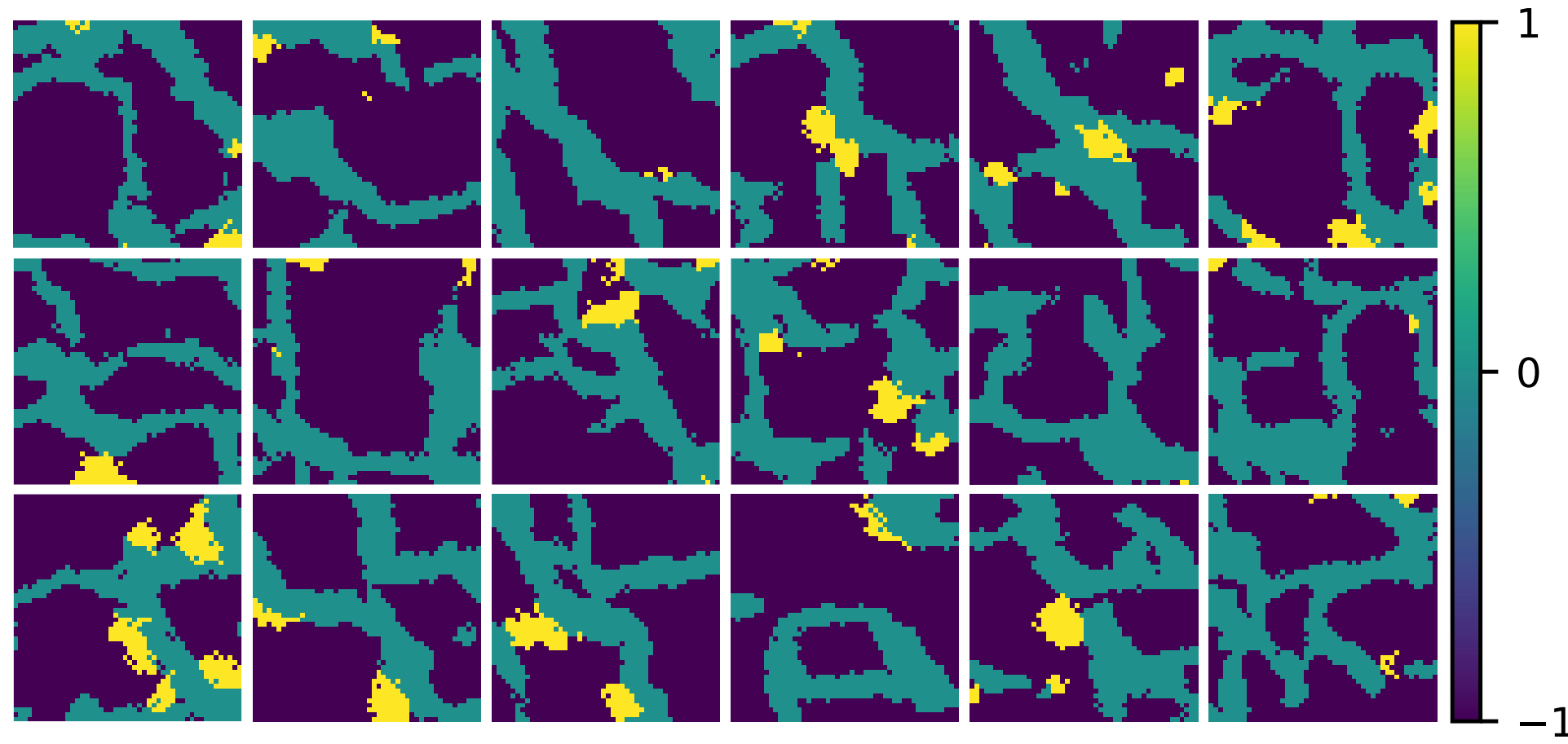

Fig. 5: Random realizations of the categorical training dataset.

This case study is a 2D model that contains 48 x 48 gridblocks with the reservoir log-permeability being the only uncertain model parameter (M = 2 304). The prior log-permeability ensemble was built using an exponential covariance function with mean equal to 3, standard deviation equal to 1.5 and correlation range equal to 10 gridblocks. Additionally, the reference model was generated using the same statistics. The reservoir simulation model contains 9 producers and 4 injectors with a configuration of 4 five-spots as we can see in Fig. 6. The producers are controlled by minimum bottom-hole pressure (BHP) equal to 200 kgf/cm$^2$ and the injectors by maximum water injection rate (WIR) equal to 500 m$^3$/day. History data consists of noisy measurements at 90 days interval in 10 measurements periods ($N_d = 660$). The covariance matrix of the measurement errors was built considering a standard deviation of 4

m$^3$/day for production rates, 3 m$^3$/day for injector rates, and 2 kgf/cm$^2$ for bottom-hole pressures.

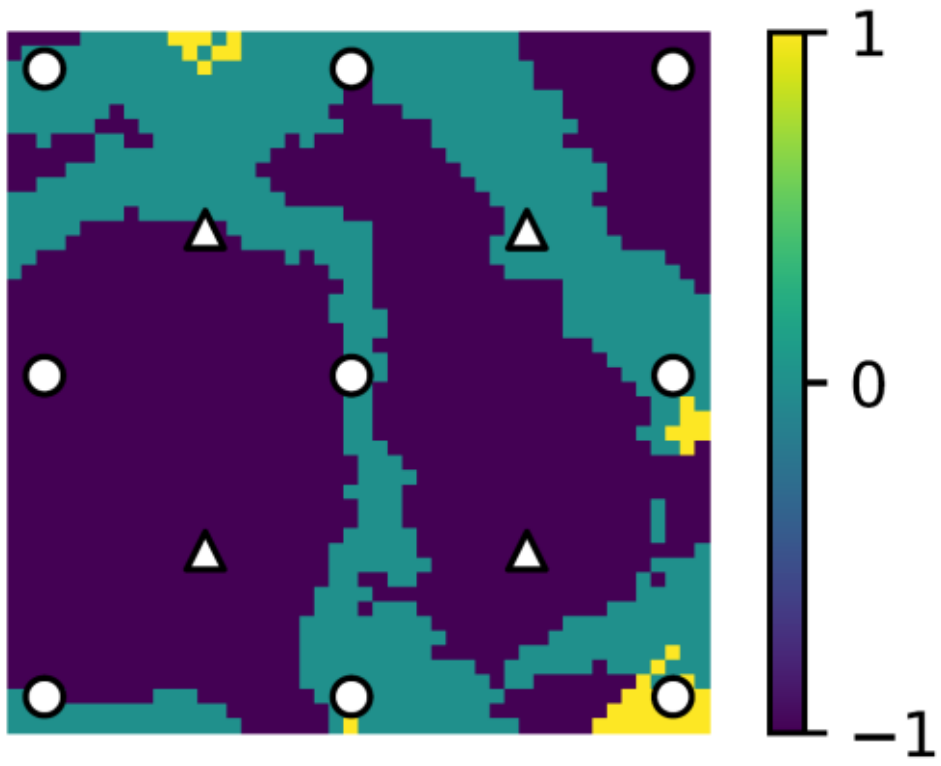

Fig. 6: The true reservoir model for the first case. The locations of injectors are indicated by triangles, and those of producers by circles.

### 3.2. Continuous Training Dataset

To create the continuous training dataset in a more realistic context, we started with the UNISIM-II-H benchmark log-permeability realizations (Correia et al., 2015; Maschio et al., 2019). The original 3D log-permeability field, which has dimensions of $46 \times 69 \times 30$, is notably non-Gaussian due to the presence of Super-K features, thin layers with exceptionally high permeability (Meyer et al., 2000; Alqam et al., 2001). Similarly to the approach used in Ranazzi et al. (2024), we generated samples by taking multiple random $48 \times 48$ crops from each of the 30 vertical layers, resulting in a total of 15 000 samples (see Fig. 7 for example of realizations).

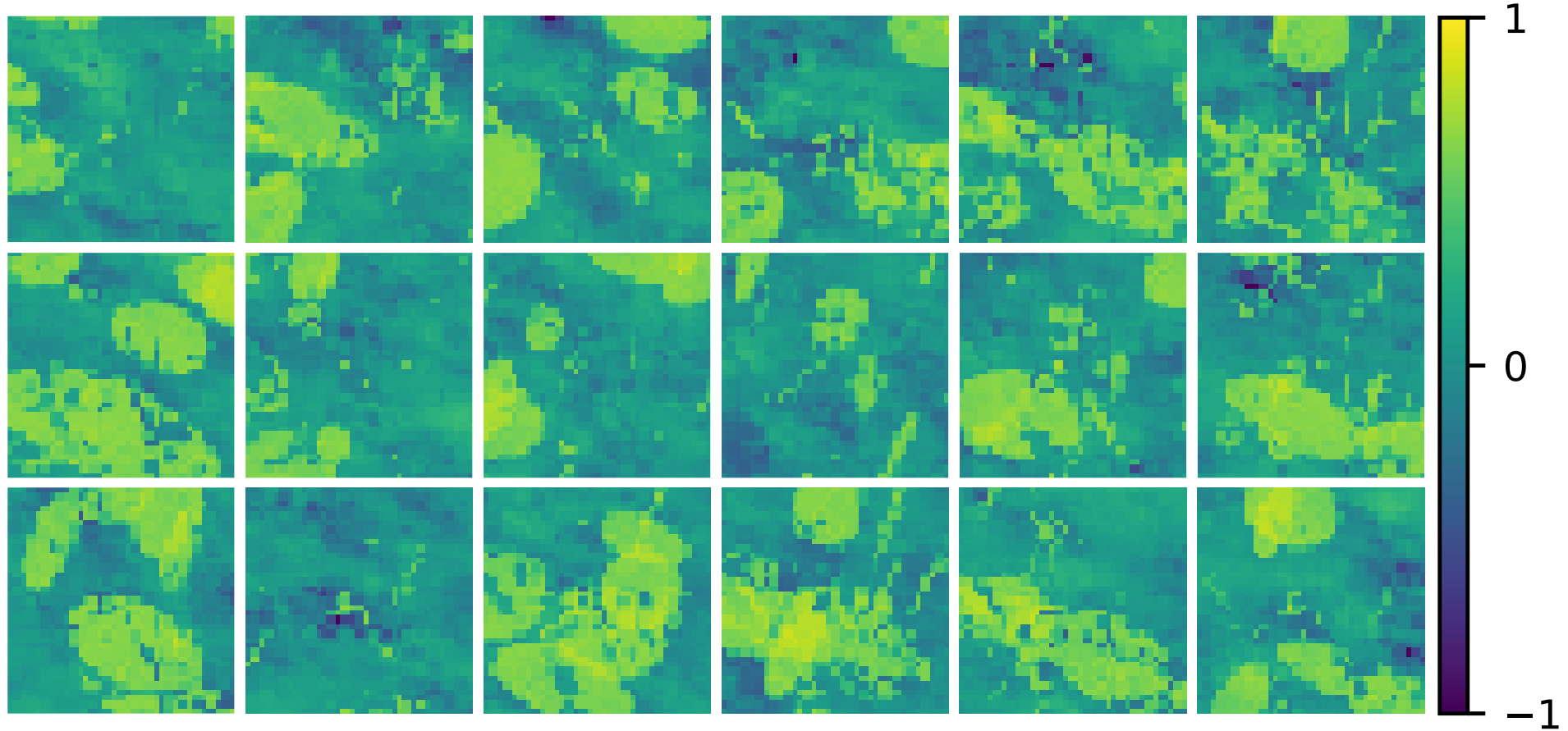

Fig. 7: Random realizations of the continuous training dataset.

This case study is a 2D model that contains 65 x 65 gridblocks with the reservoir log-permeability being the only uncertain model parameter (M = 4 225). The prior log-permeability ensemble was built using an exponential covariance function with mean equal to 3, standard deviation equal to 1.5 and correlation range equal to 10 gridblocks. Additionally, the reference model was generated using the same statistics. The reservoir simulation model contains 9 producers and 4 injectors with a configuration of 4 five-spots. The well control strategy is identical to that adopted in the previous case study. History data consists of noisy measurements at 90 days interval in 10 different measurements periods ($N_d = 660$). The covariance matrix of the measurement errors was built considering a standard deviation of 4 $m^3$/day for production rates, 3 $m^3$/day for injector rates, and 2 kgf/$cm^2$ for bottom-hole pressures. The true reservoir model with well positions is shown in Fig. 8.

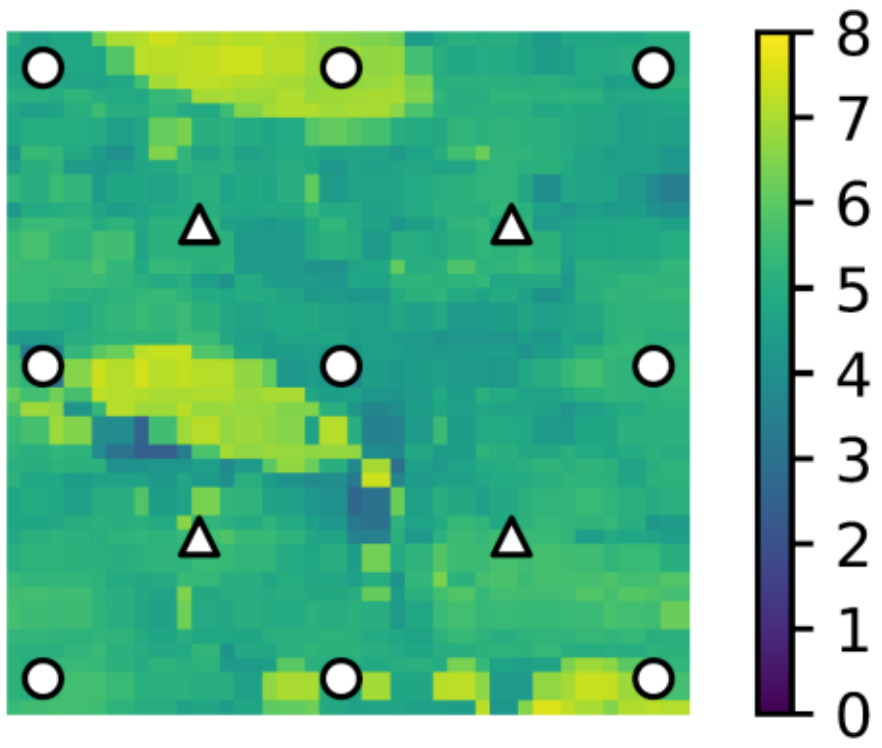

Fig. 8: The true reservoir model for the second case. The locations of injectors are indicated by triangles, and those of producers by circles.

### 3.3. DCGAN Structure and Configuration

The generator progressively upsamples a 512-dimensional latent vector to high resolution using upsampling blocks, consisting of convolutional and resize layers, plus batch normalization and ReLU activations. Conversely, the discriminator utilizes a symmetric downsampling architecture with convolutional and average pooling layers, plus Leaky ReLU activations. Training employed a non-saturating GAN loss with R1 regularization, with Adam optimizer with a learning rate of 0.0001 and fixed batch size of 32. Training was carried out until 150 000 iterations for categorical dataset and 200 000 for continuous dataset. To prevent overfitting, early stopping was triggered after 20 metric measurement of stagnant FID scores. DCGAN configuration details are provided in the Appendix.

### 3.4. DCVAE Structure and Configuration

The encoder compresses input using strided convolutions with 32, 64, 128, and 256 filters. Leaky ReLU (slope 0.2) and batch normalization layers are applied at each convolutional layer. The architecture culminates in two parallel dense layers that parameterize a 512-dimensional Gaussian distribution—one outputting the mean vector and another producing the logarithmic variance. The decoder network executes the inverse transformation, mapping the latent vector back to the original data space.

The model was trained using the Adam optimizer (Learning rate of 0.001) for up 200 epochs with a batch size of 256. A critical aspect of the VAE formulation is the weighting factor $\beta$, set to 0.05, which balances the reconstruction fidelity against the regularization imposed by the Kullback-Leibler divergence. To ensure stability, the learning rate was reduced by 0.5 after 5 epochs of validation loss stagnation. Early stopping was employed with a patience of 20 epochs and a minimum delta of 0.0005. Further DCVAE configuration details are in the Appendix.

### 3.5. VAE-GAN Structure and Configuration

The encoder network processes through a series of convolutional layers with filter counts specifically designed to create an information bottleneck: 58 filters in the initial layer, expanding to 116, and finally 230 filters, each employing a kernel size of 5 with stride-2 downsampling. This carefully calibrated progression ensures gradual abstraction of spatial features while maintaining sufficient representational capacity for complex geological patterns. The encoder culminates in two parallel dense layers parameterizing a 512-dimensional Gaussian latent distribution, with the sampling operation implementing the reparameterization trick to enable gradient propagation through stochastic nodes. Beginning with a dense layer, the generator performs progressive upsampling through transposed convolutional layers that mirror the encoder's architecture in reverse. This symmetrical design ensures consistency between encoding and decoding pathways while maintaining the capacity to reconstruct input data faithfully. The discriminator network employs spectral normalization on all convolutional and dense layers, a critical stabilization technique that prevents mode collapse and training instability commonly observed in GAN architectures. Gaussian noise injection at the discriminator input further enhances training robustness by preventing overconfidence in classification decisions.

The training process combines four distinct loss components, each contributing to different aspects of model performance. The reconstruction loss employs both L1 and L2 norms, with L1 promoting edge preservation and L2 ensuring overall structural fidelity. The Kullback-Leibler divergence with a weighting factor $\beta = 0.2$ encourages the latent space to approximate a standard normal distribution while preserving sufficient expressiveness for accurate reconstruction. The perceptual loss, weighted by $\gamma = 0.1$, extracts feature from both InceptionV3 and a pre-trained reservoir classifier, enforcing similarity in semantic feature spaces beyond pixel-level comparisons. The training regimen employs separate Adam optimizers for generator and discriminator components, each with learning rate decay scheduling and gradient clipping at norm 1.0 to ensure stable convergence. The configuration of VAE-GAN is summarized in the Appendix.

## 4. RESULTS AND DISCUSSION

For each case study, we present the results of training of the models followed by the results of data assimilation. All training process were performed in the same standard Nvidia GPU (GeForce RTX 3060 with 16 GB of memory) of a stand-alone computer, using the machine learning framework TensorFlow 2.10 (Abadi et al., 2015).

### 4.1. Case Study 1: Categorical training dataset

#### *4.1.1. Training of Models*

***Deep Convolutional Generative Adversarial Network (DCGAN)***

All models were trained using the categorical training dataset with all 80 000 samples for training until reach early stopping criteria. This large number of samples was used to ensure that sample size was not a limiting factor for our models. Determining the sufficient number of samples for each model would require a separate study. The need for a large number of samples can be a limiting factor in practical applications, but recent study (Ranazzi et al., 2024) show that techniques such as data augmentation can be used to generate a large number of samples for training deep learning models. We trained the DCGAN model after finding the better structure and configuration. The training time was 4 hours and 8 minutes. We obtained the FID and FRD after the end of iterations, respectively equal to 203.36 and 0.23, showing that the training was efficient and FRD a more appropriate metric for this case study. This can also be proven by the high-quality of the images generated at the end of the training, as shown in Fig. 9.

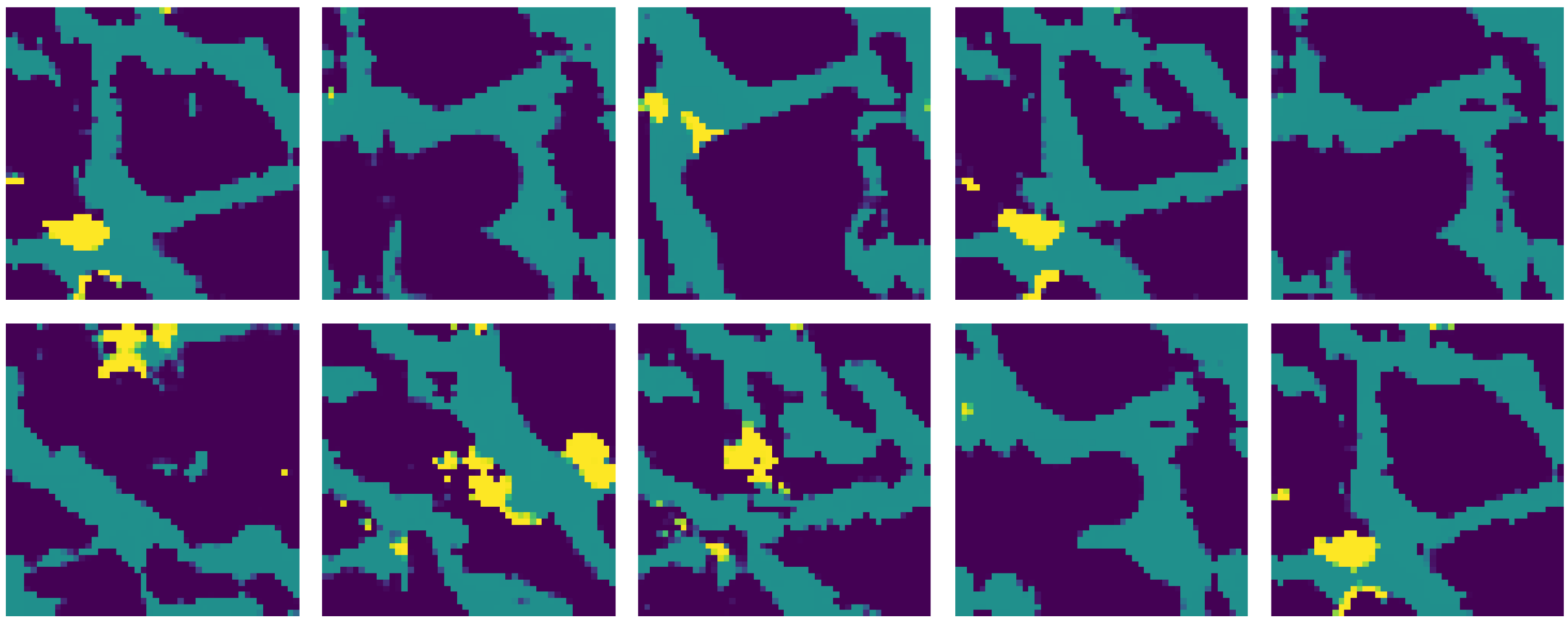

Fig. 9: Images generated by DCGAN after training with 150 000 iterations for case study 1.

Fig. 10 shows the original and generated distributions, showing that DCGAN tends to follow the original distribution of the data (only with three values).

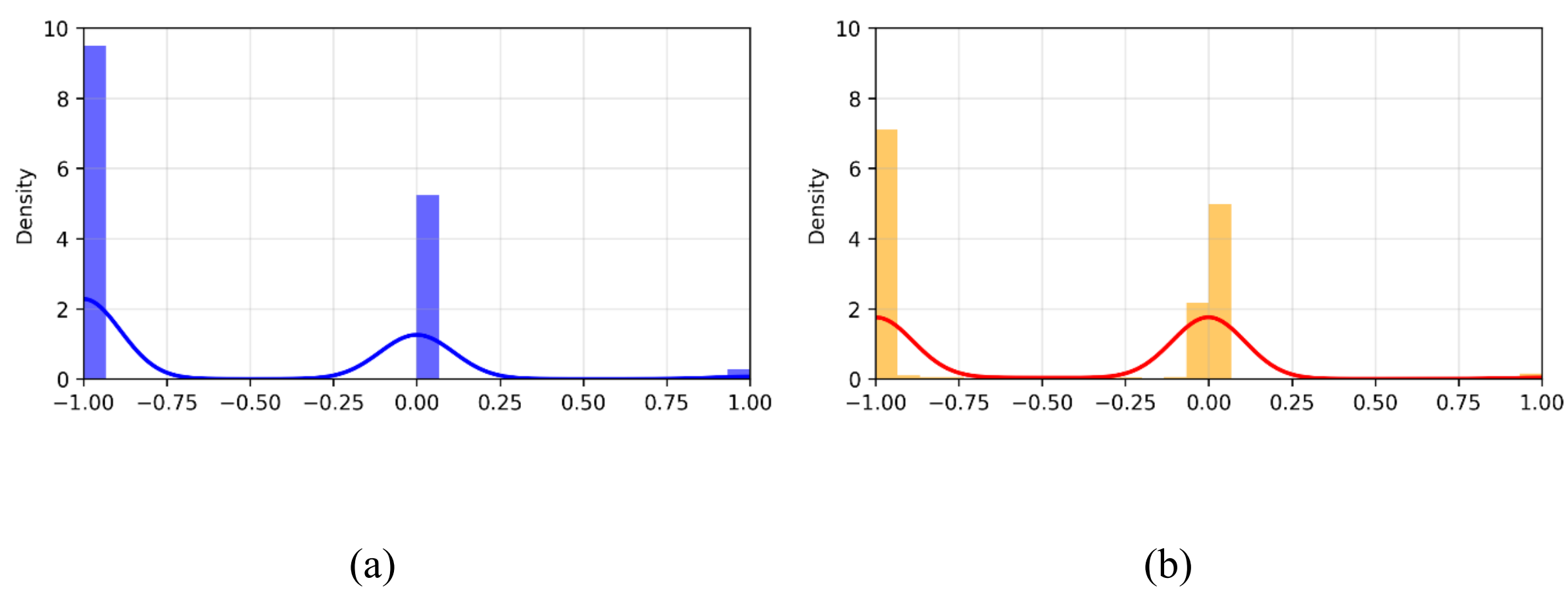

Fig. 10: Comparison between distributions of (a) original and (b) generated data for DCGAN in the categorical case after training.

### *Deep Convolutional Variational Autoencoder (DCVAE)*

The DCVAE model was trained until reach early stopping criteria, reaching the FID and FRD after the end of iterations, respectively equal to 1209 and 37.4. The training time was 41 minutes, showing that the computational cost of training VAE is much lower than DCGAN, about 6 times less time. We can see in Fig. 11 the quality of the images reconstructed by the decoder's model, showing that the images were blurry and with low quality, as expected from literature.

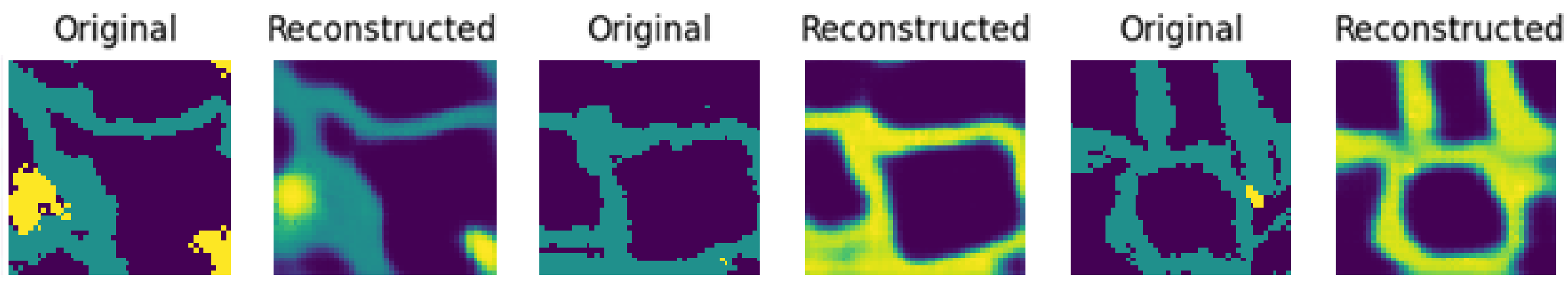

Fig. 11: Three examples of random images from the set along with the reconstructed images by DCVAE for case study 1.

Fig. 12 show the decrease in the reconstruction loss and Kullback-Leibler loss curves throughout the training. We can see that the reconstruction loss was always decreasing, tending to stabilize around 40 epochs and that the KL loss stabilizes after 25 epochs with low errors. This shows that DCVAE training was efficient and sufficient.

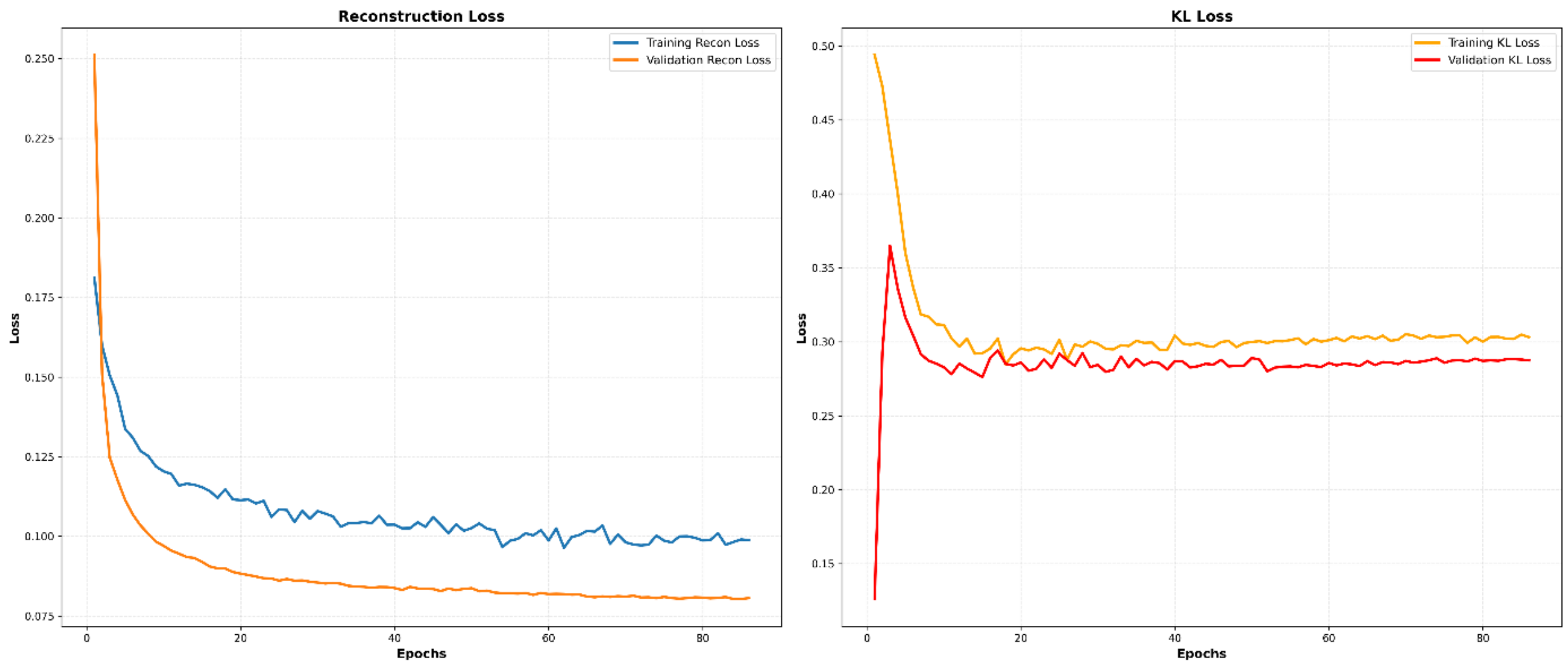

Fig. 12: Graphs of reconstruction (left) and Kullback-Leibler loss (right) throughout the training with DCVAE for case study 1.

Fig. 13 shows the evolution of the MSE of the training and validation sets, showing that the training and validation error curves stabilized after 40 epochs.

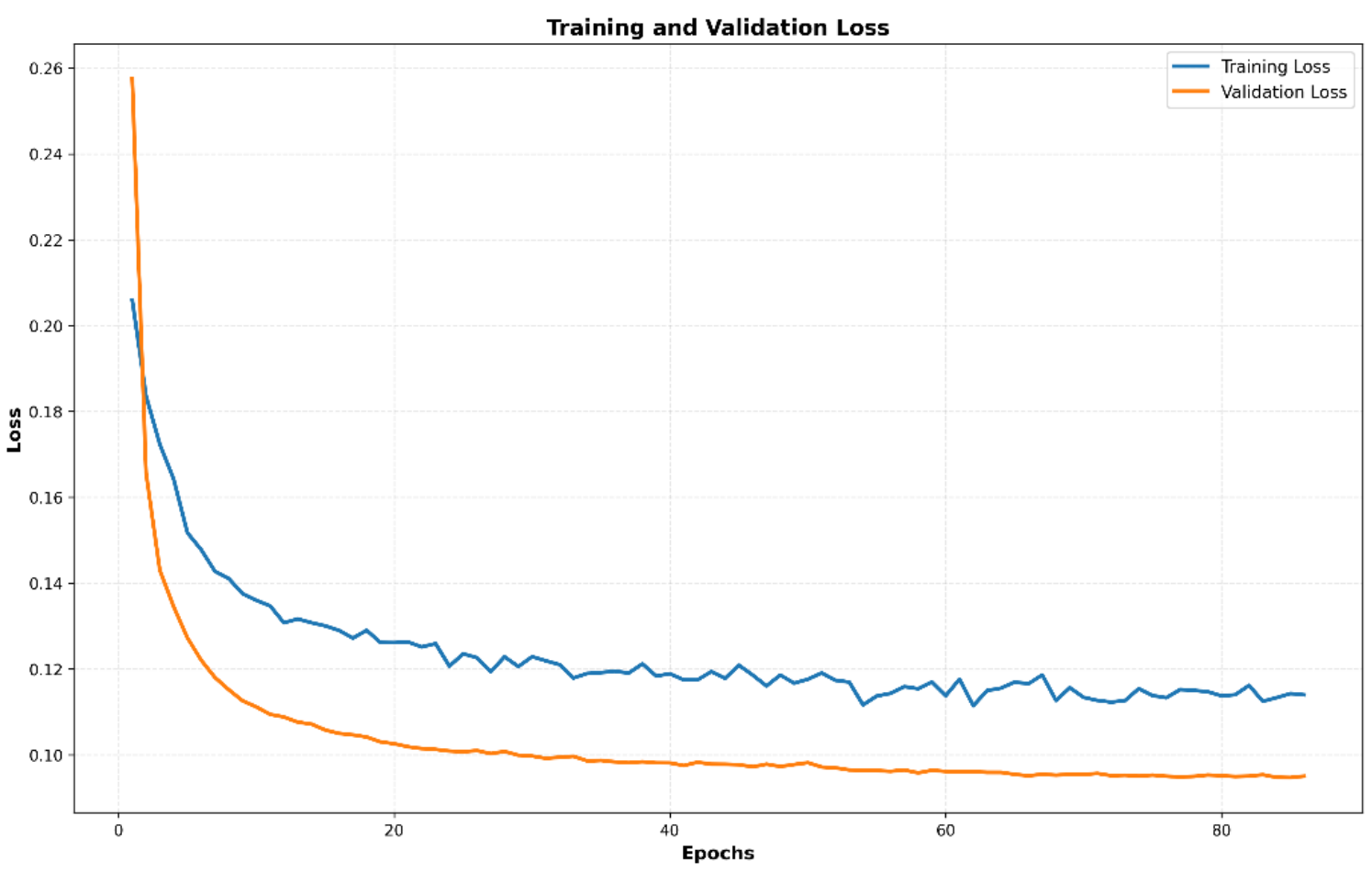

Fig. 13: MSE of training and validation versus epochs with DCVAE across training iterations.

Fig. 14 shows the original and latent space distributions, showing the transformation of the initial distribution (non-Gaussian) into a Gaussian distribution in the latent space. Here we can see that this type of model is very efficient in transforming a non-Gaussian distribution into a Gaussian one, which ends up reflecting in better data assimilation in relation to GANs, as we will see later in this work and as has already been observed in works in the literature.

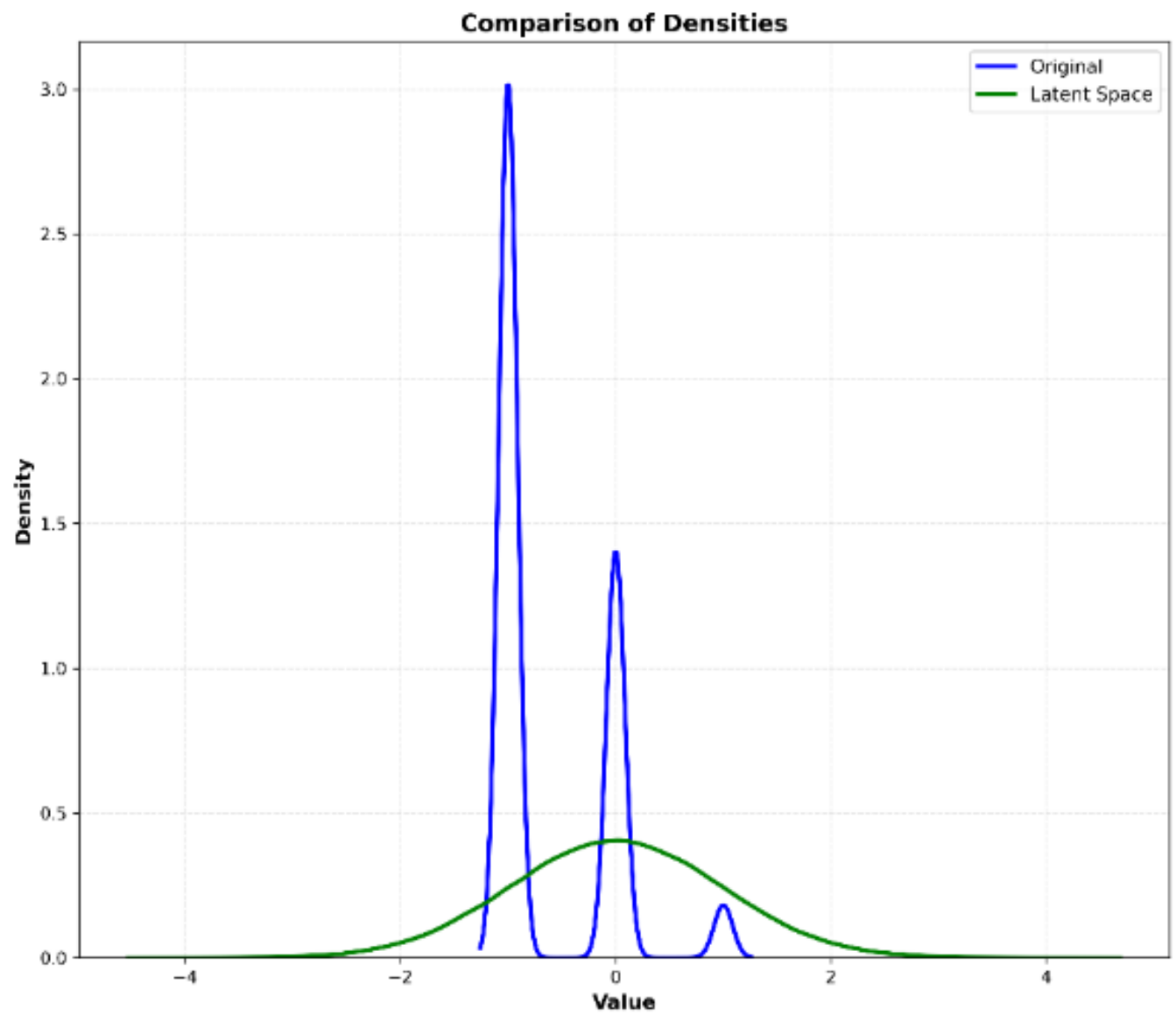

Fig. 14: Comparison between distributions of original data and latent space for VAE in the case study 1.

***Variational Autoencoder Generative Adversarial Network (VAE-GAN)***

The VAE-GAN model was trained until reach early stopping criteria, reaching the FID and FRD after the end of iterations, respectively equal to 382 and 5.12. The training time was 1 hour and 47 minutes, proving to be slightly more computationally expensive than the VAE model, but significantly less so than the GAN. We can see in Fig. 15 the high-quality of the images reconstructed by the decoder's model, showing images with higher quality than VAE, and with similar quality to DCGAN.

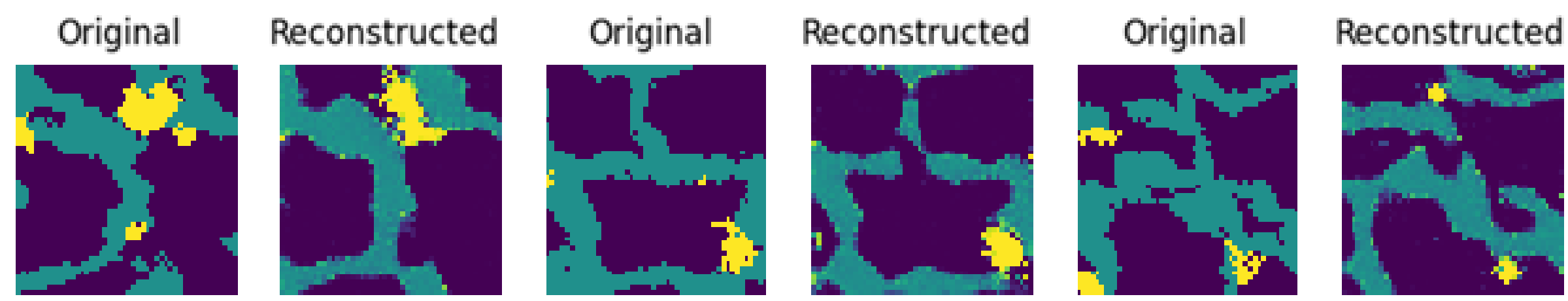

Fig. 15: Three examples of random images from the set along with the reconstructed images by VAE-GAN for case study 1.

Fig. 16 show the reconstruction and Kullback-Leibler losses throughout the training. As we can see, the reconstruction loss dropped rapidly at first, stabilizing after 20 epochs. The KL loss, on the other hand, has an initial rise, followed by a fall, and then stabilizes after 20 epochs.

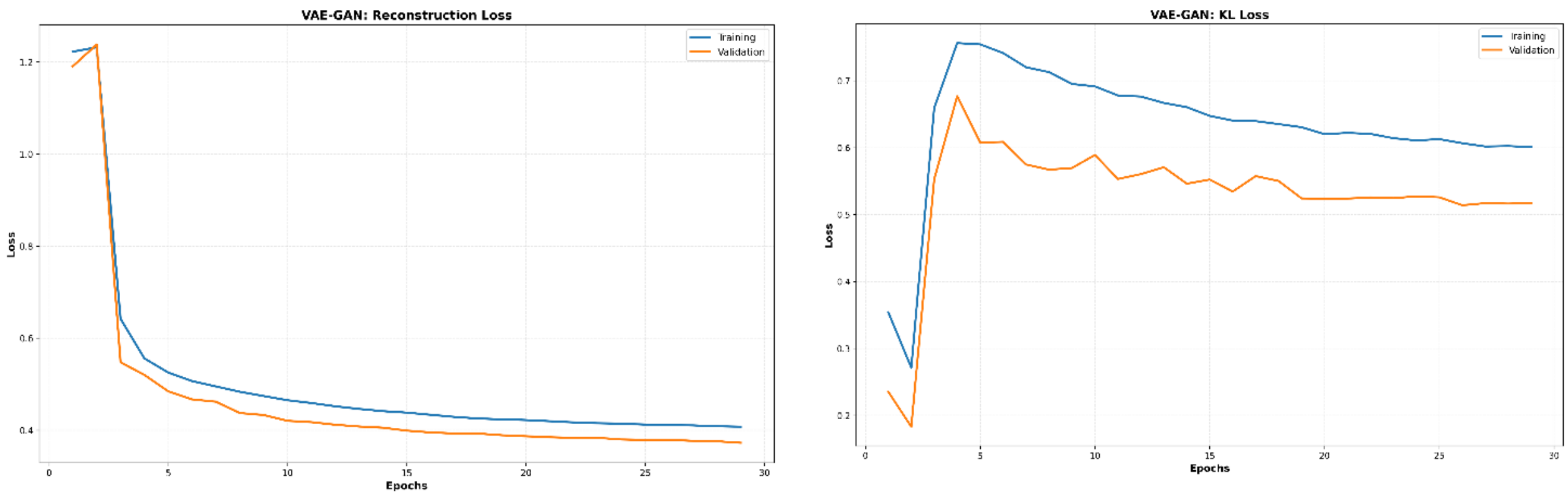

Fig. 16: Reconstruction and Kullback-Leibler losses throughout the training with VAE-GAN in the categorical case.

Fig. 17 shows the total error for the training and the validation sets. This graph is the combination of all losses of this model: reconstruction, KL, generator and discriminator losses. The curves show a stabilization in errors after 15 epochs. We can observe initial instability until stability is reached, since it is necessary to train three networks at the same time: encoder, decoder and discriminator.

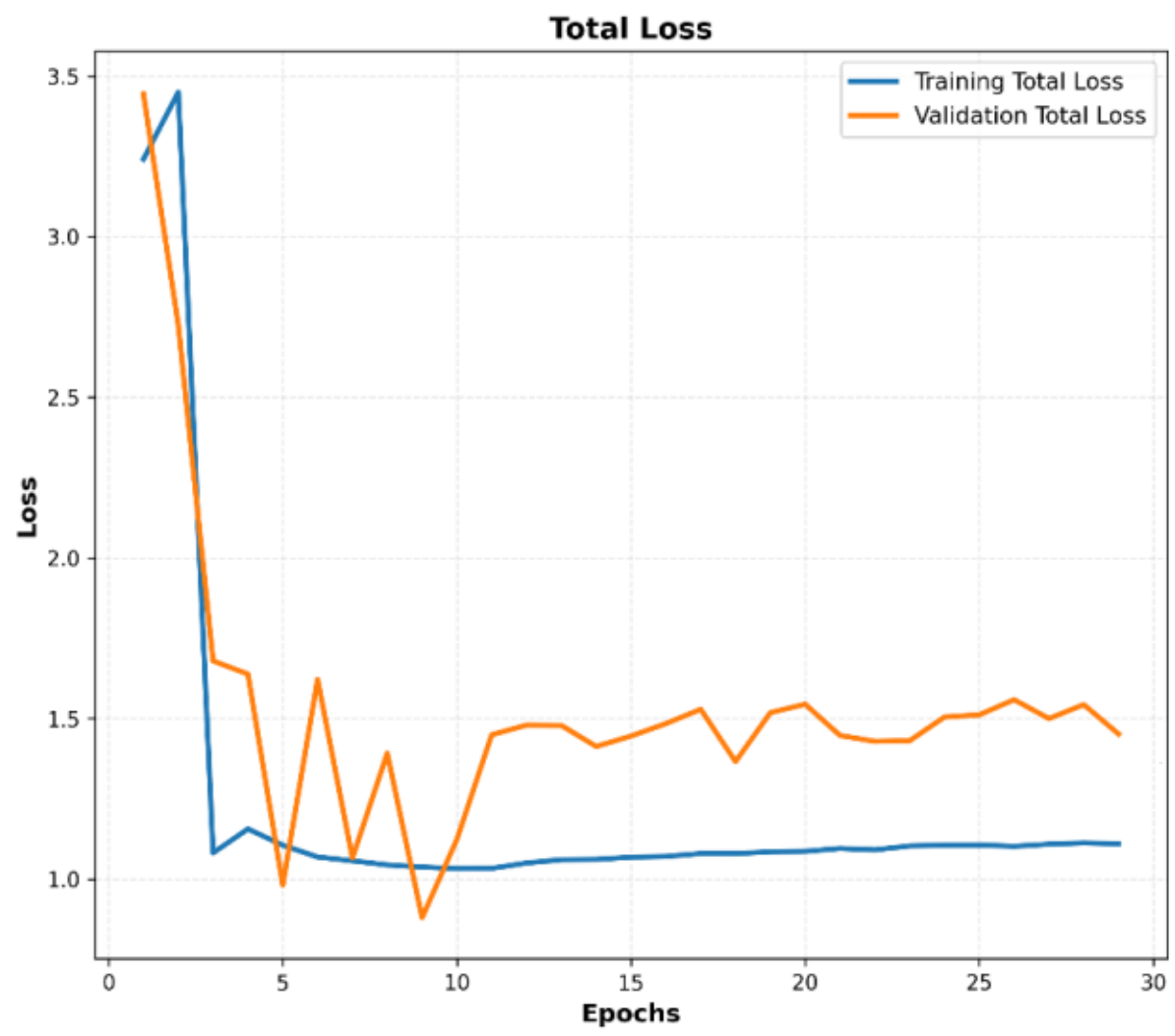

Fig. 17: Total loss versus epochs with VAE-GAN in the categorical case.

Fig. 18 shows the original and latent space distributions, showing the transformation of the initial distribution into a Gaussian distribution. Here we can see that, like VAE, the VAE-GAN model is also very efficient in transforming a non-Gaussian distribution into a Gaussian one.

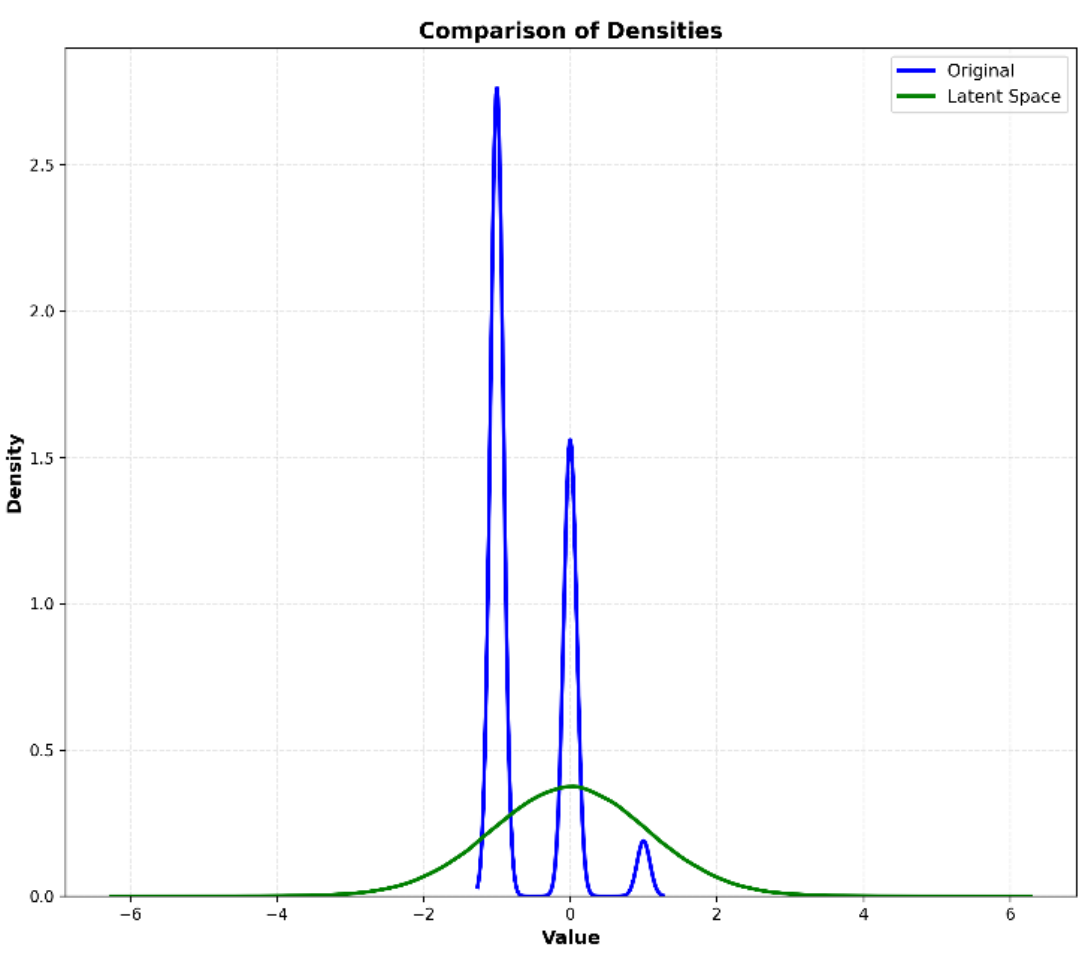

Fig. 18: Comparison between distributions of original data and latent space for VAE-GAN in the categorical case.

In the Table 1, we summarize the results of the FID and FRD metrics at the end of training for the three models. We can conclude again that the FRD metric was more appropriate than the FID for our case study and that the values obtained were low, demonstrating that the training of all models was efficient. It is interesting to note that the hybrid model (VAE-GAN) achieved an FRD value intermediate between GAN and VAE.

Table 1: Results of FID and FRD at the end of training of models in the categorical case.

| | FID | FRD |
|---|---|---|
| DCGAN | 203.36 | 0.23 |
| DCVAE | 1209 | 37.40 |
| VAE-GAN | 382 | 5.12 |

#### *4.1.2. Data Assimilation*

For ESMDA, we updated directly the natural logarithm of permeability. In Fig. 19, we can see the images of the true case, the priori and posteriori mean and standard deviation. We can observe that the posterior mean and standard deviation present an image like the true case in the cases of VAE and VAE-GAN. This shows that the VAE and VAE-GAN models were able to preserve the variance and mean values of the initial ensemble more than GAN model.

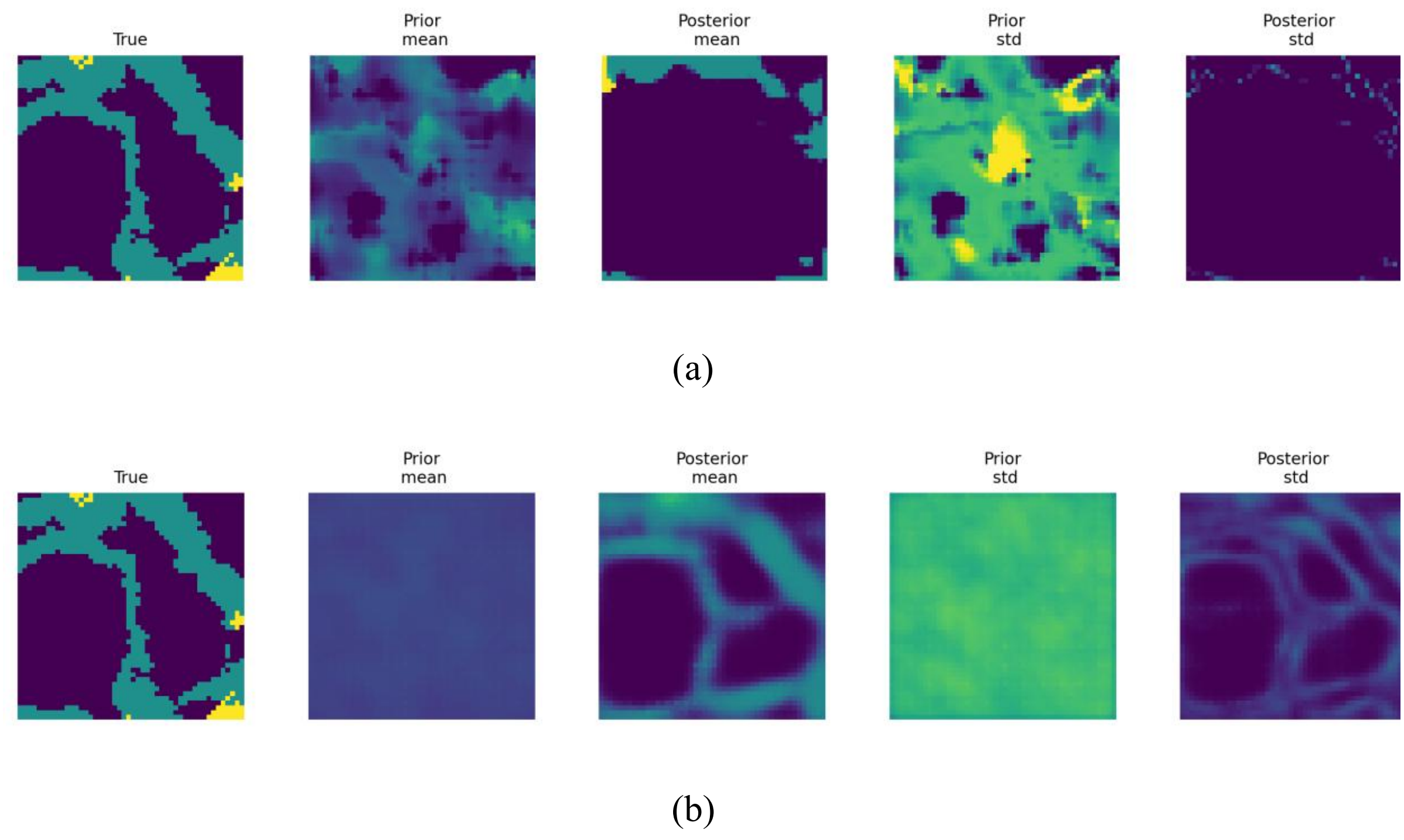

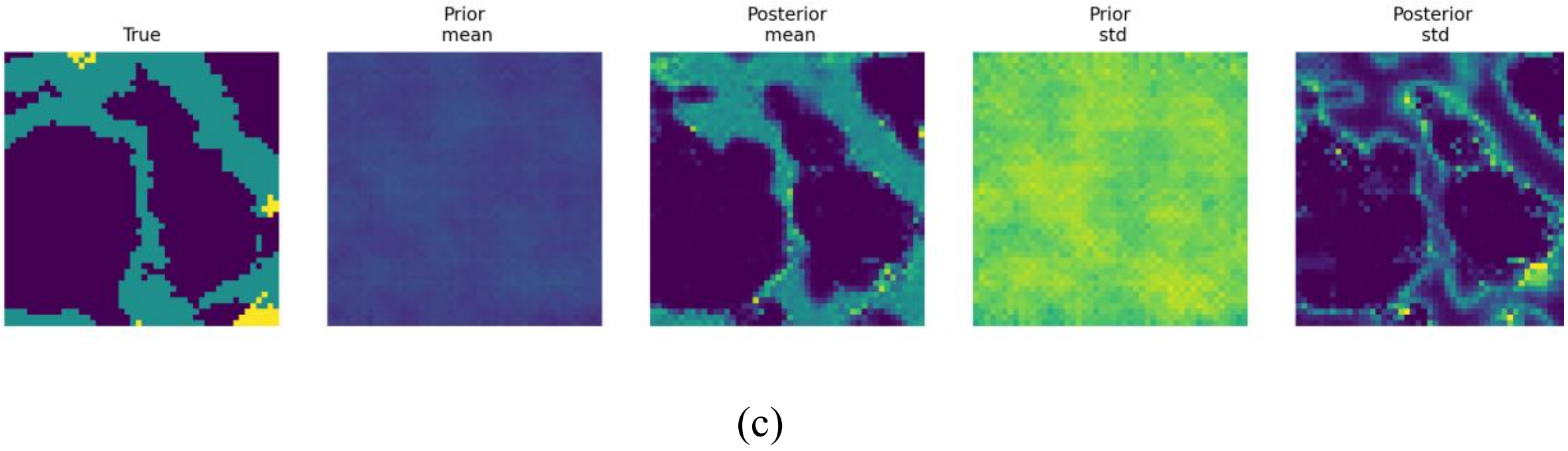

(c)

Fig. 19: Images of the true case, the priori and posteriori mean, and the priori and posteriori standard deviation in the categorical case for: (a) DCGAN, (b) DCVAE and (c) VAE-GAN.

In Fig. 20, we can see the images of the priori and posteriori of member 1, as an example, to illustrate a case before and after assimilation. As we can see, the GAN network fails to preserve the features of member 1. In VAE, the network can delineate the model's features, but with low resolution. In the VAE-GAN model, the network can delineate the facies contours well and with better resolution than the VAE network.

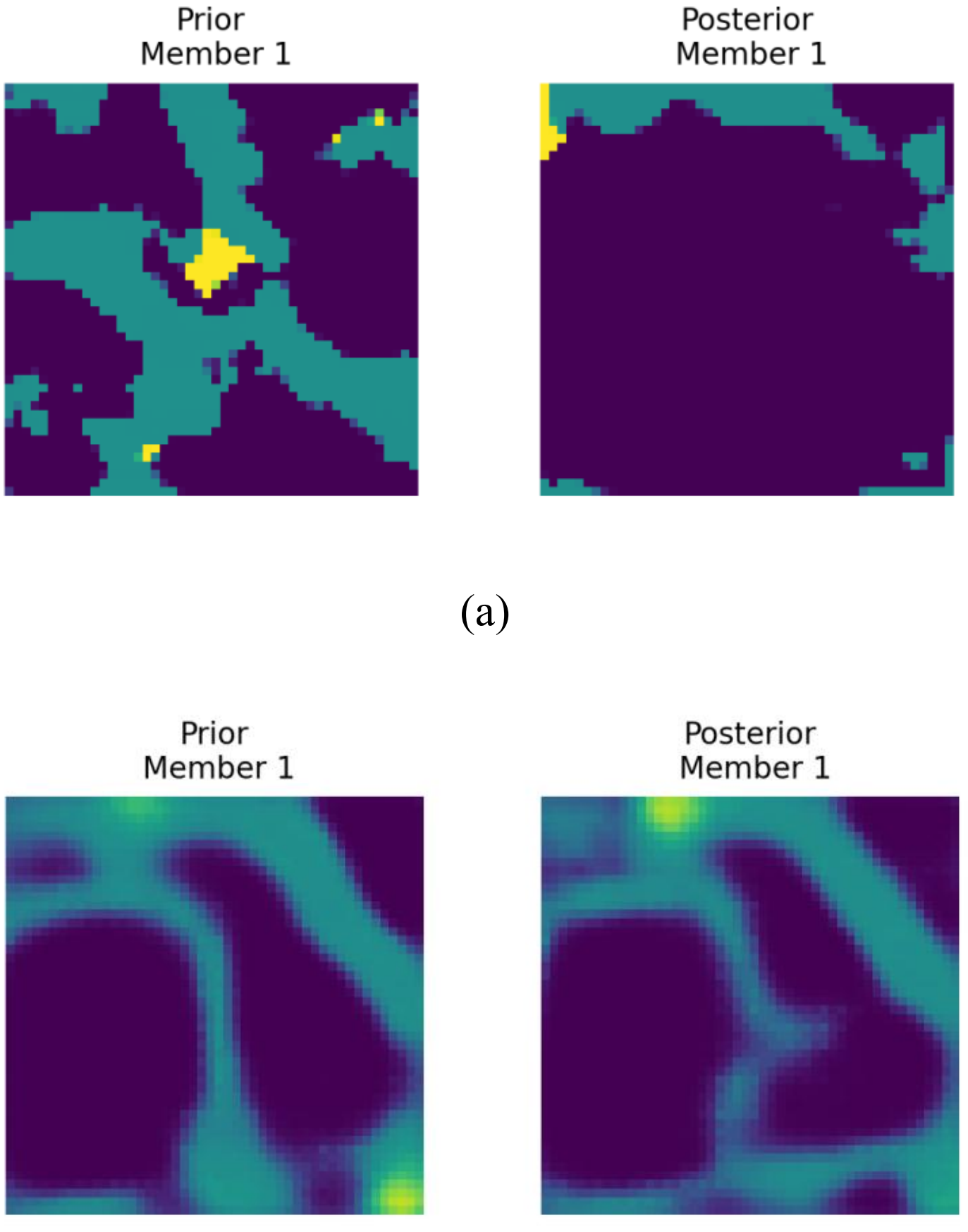

(a)

(b)

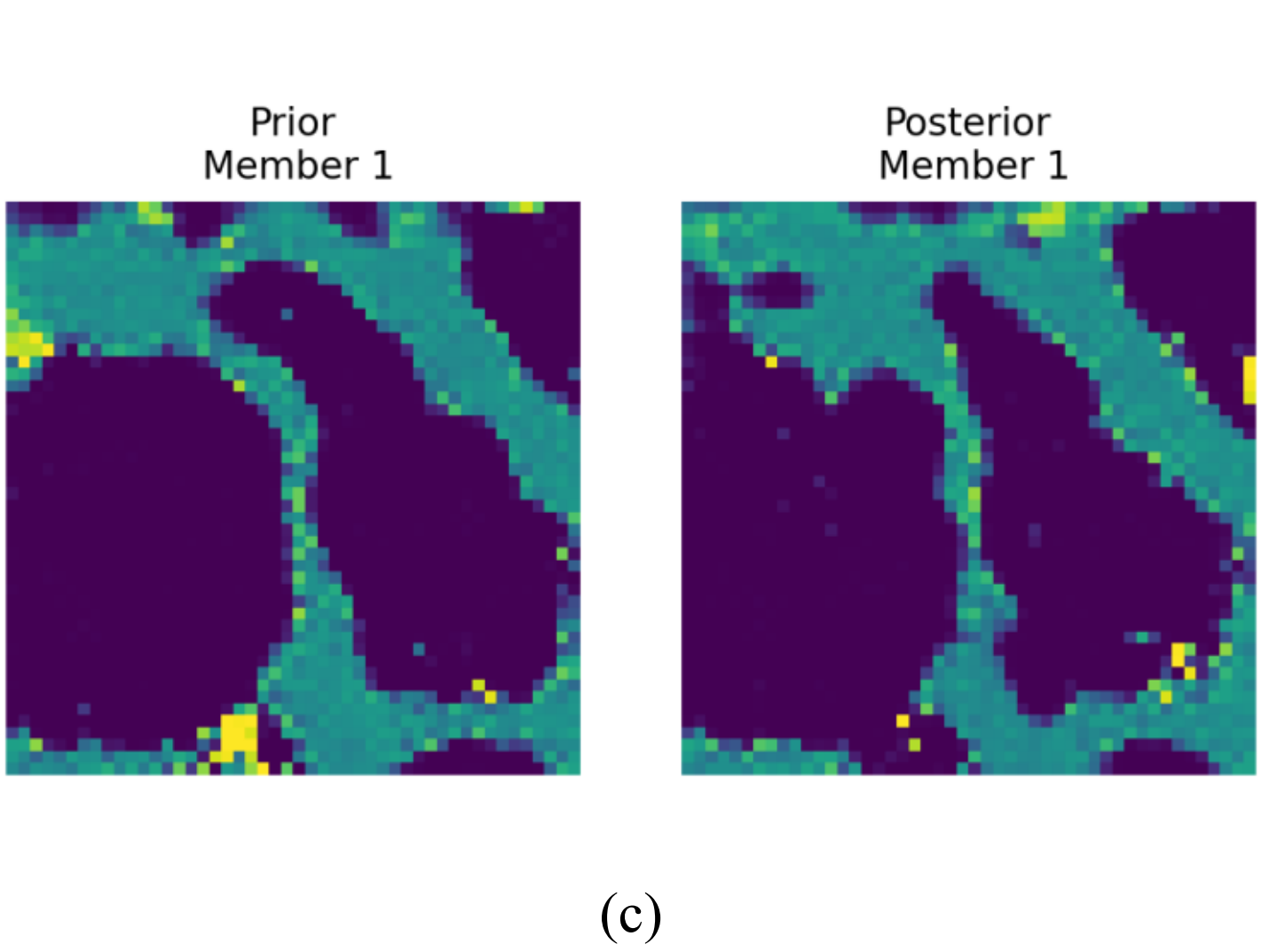

(c)

Fig. 20: Images of the priori and posteriori of member 1 in the categorical case for: (a) DCGAN, (b) DCVAE and (c) VAE-GAN.

Analyzing the time series of production data for P1, P2, P3, P4, P5 and P6 wells in Fig. 21, it is possible to confirm that no ensemble collapse occurred. In addition, all cases resulted in reductions in terms of ensemble spread. We can also observe that the GAN model presented the worst matching for all wells. For the VAE model, we can observe a good matching for all wells. For the VAE-GAN model, we can observe a good matching for P2, P3 and P5 wells and worse matchings for P1 and P5, for oil and water production as a function of time.

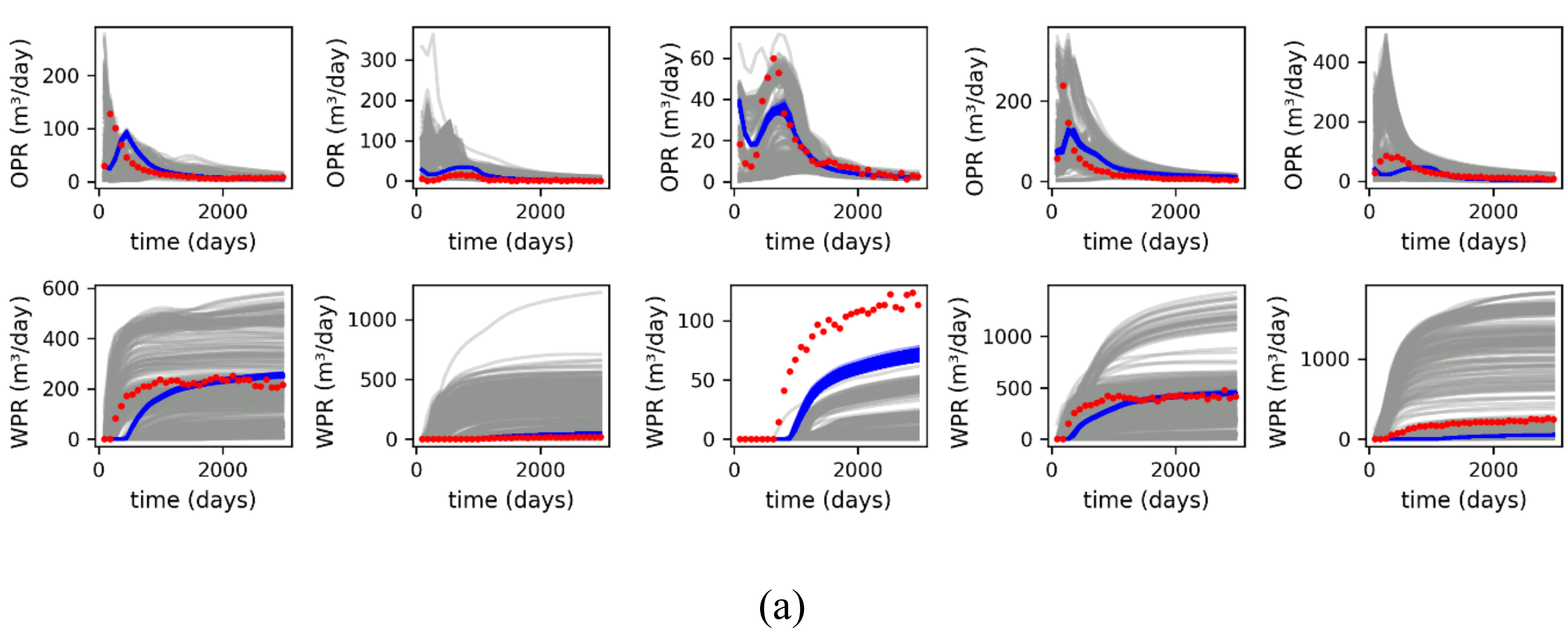

(a)

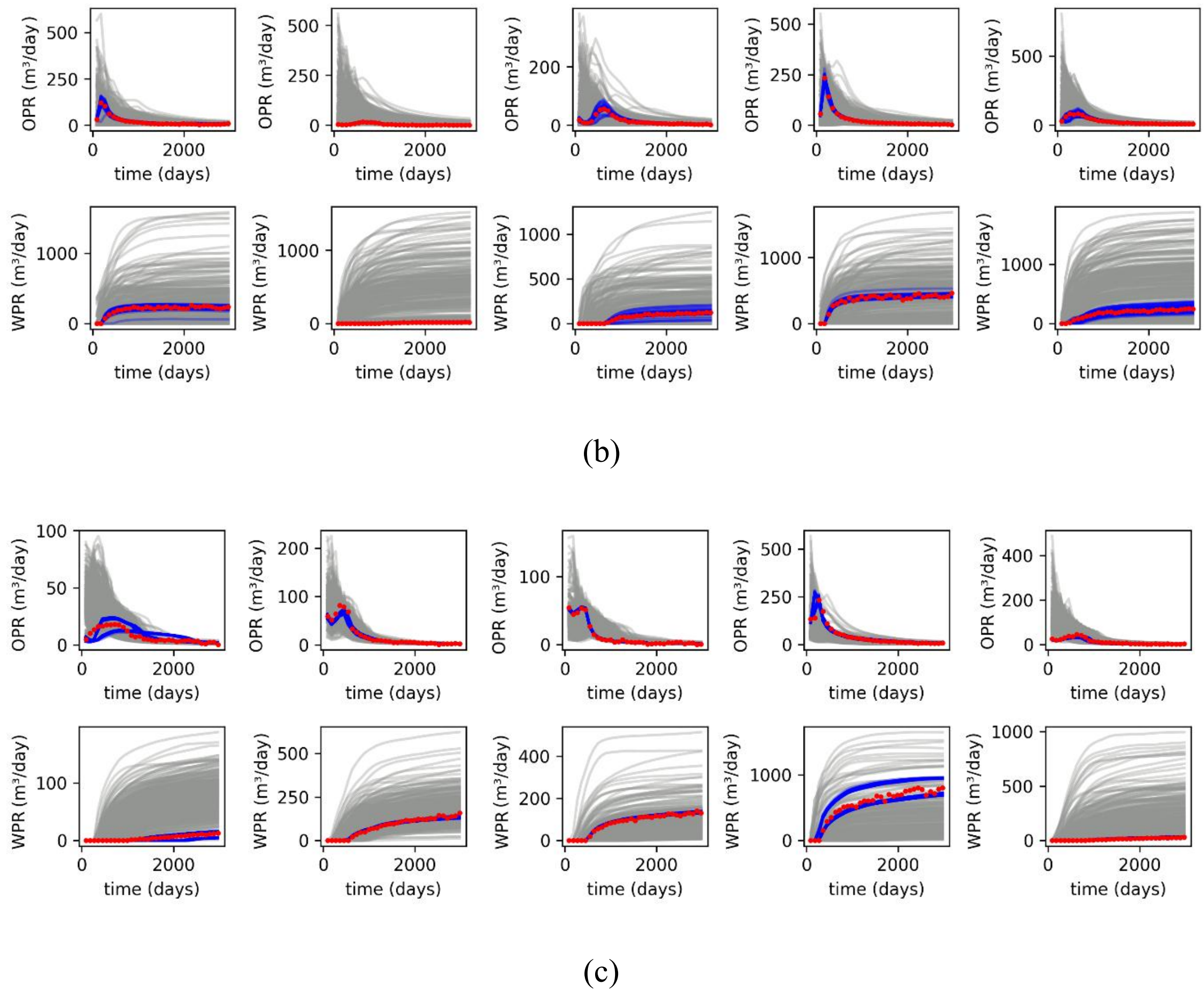

Fig. 21: Time series of production data from the first five producers of oil production rate (above) and water production rate (below). Here, the gray lines represent the prior ensemble, blue lines represent the posterior ensemble, and the red dots represent the measurements in the case study 1 for: (a) DCGAN, (b) DCVAE and (c) VAE-GAN.

Fig. 22 shows graphs of data mismatching, RMSE, spread and balanced accuracy as a function of the number of iterations. As we can see, there was an increasingly better matching when moving from GAN to VAE and VAE-GAN models. The results showed better values for VAE and VAE-GAN for RMSE, spread, data mismatching (DM) and balanced accuracy than GAN model.

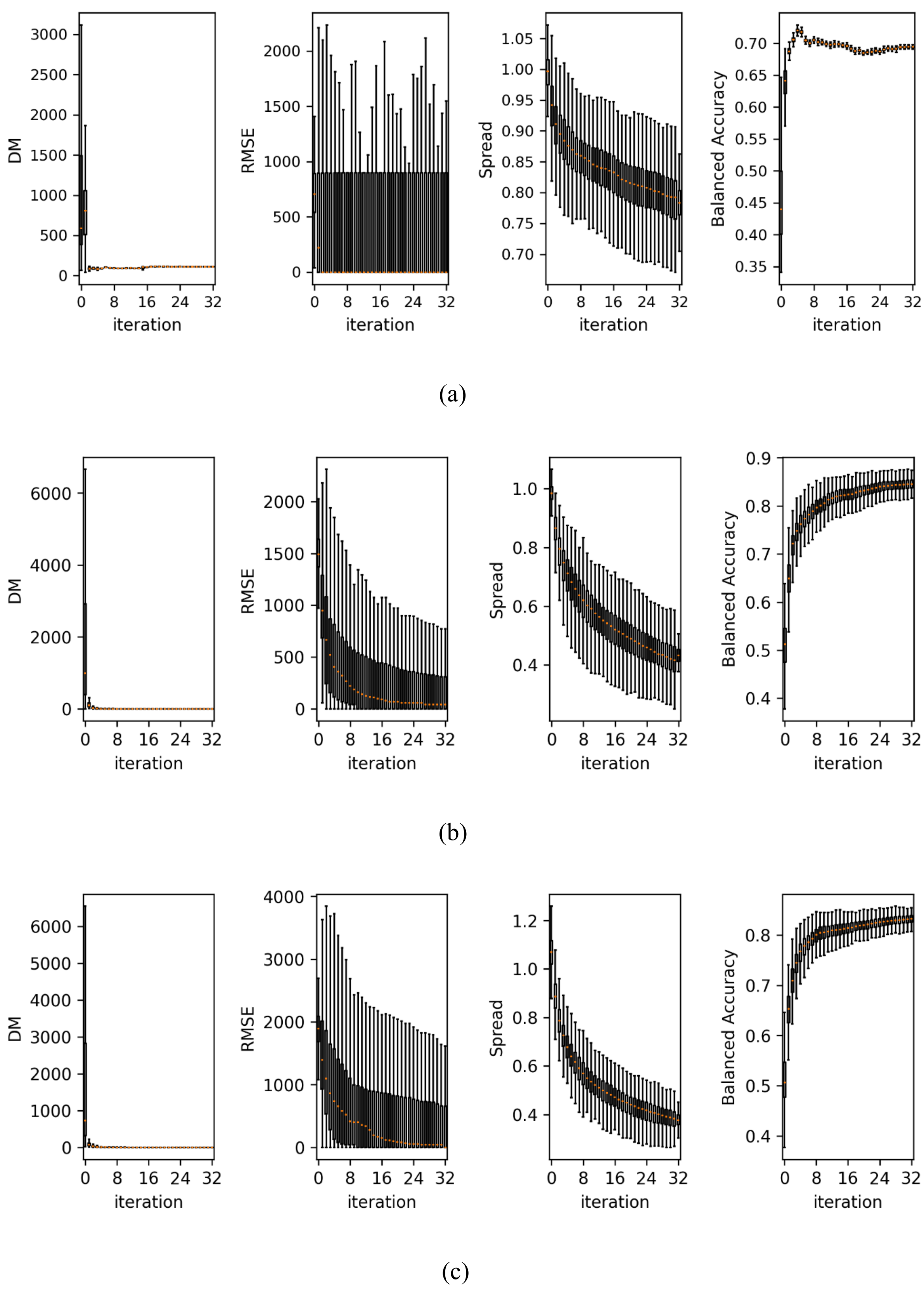

Fig. 22: Graphs of data mismatching, RMSE, spread and balanced accuracy as a function of the number of iterations in the categorical case for: (a) DCGAN, (b) DCVAE and (c) VAE-GAN.

## 4.2. Case Study 2: Continuous training dataset

### *4.2.1. Training of Models*

***Deep Convolutional Generative Adversarial Network (DCGAN)***

We trained the DCGAN model with continuous case after found the better structure and configuration using the categorical training dataset with all 50 000 samples for training until reach the early stopping criteria. The training time was 3 hours and 58 minutes. We obtained the FID and FRD after the end of iterations, respectively equal to 32 and 0.26, showing that training was enough. This can be seen by the high-quality of the images generated at the end of the training, as shown in Fig. 23.

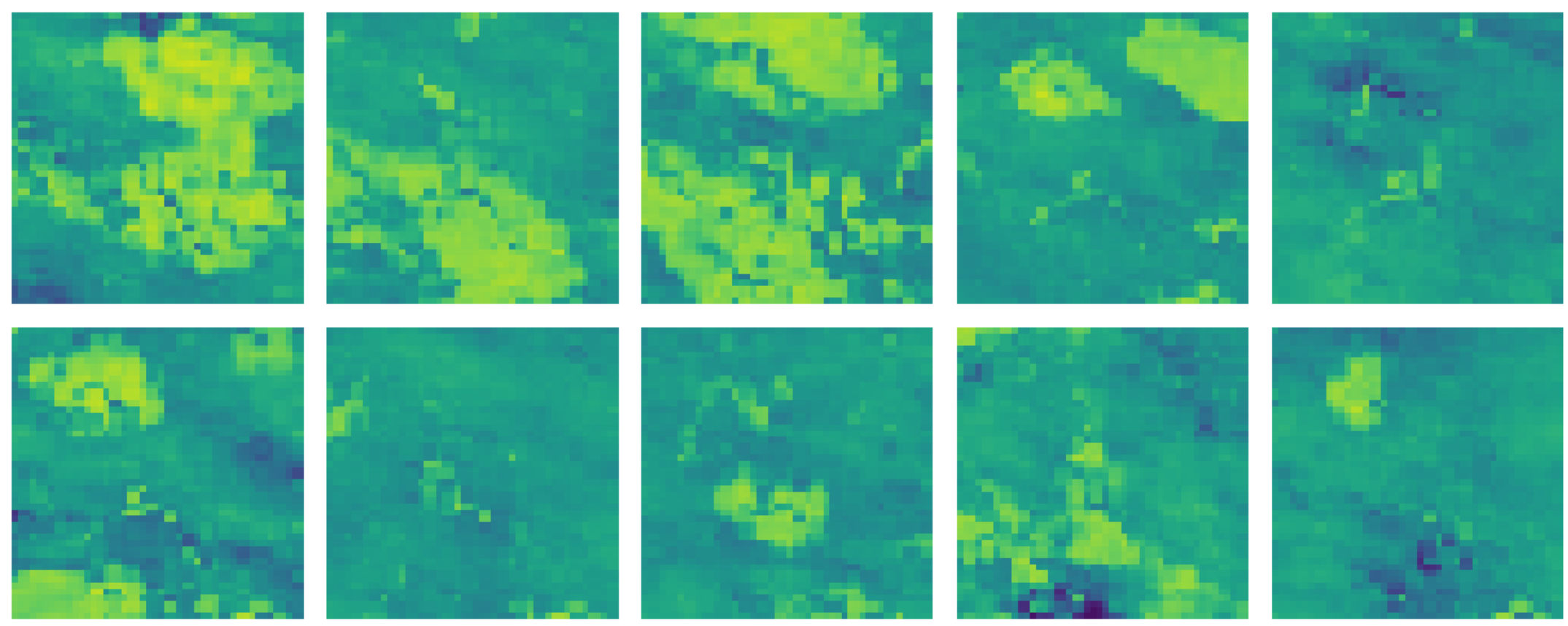

Fig. 23: Images generated by DCGAN after training in the continuous case.

Fig. 24 shows the original and generated distributions, showing that DCGAN tends to follow the original distribution of the data.

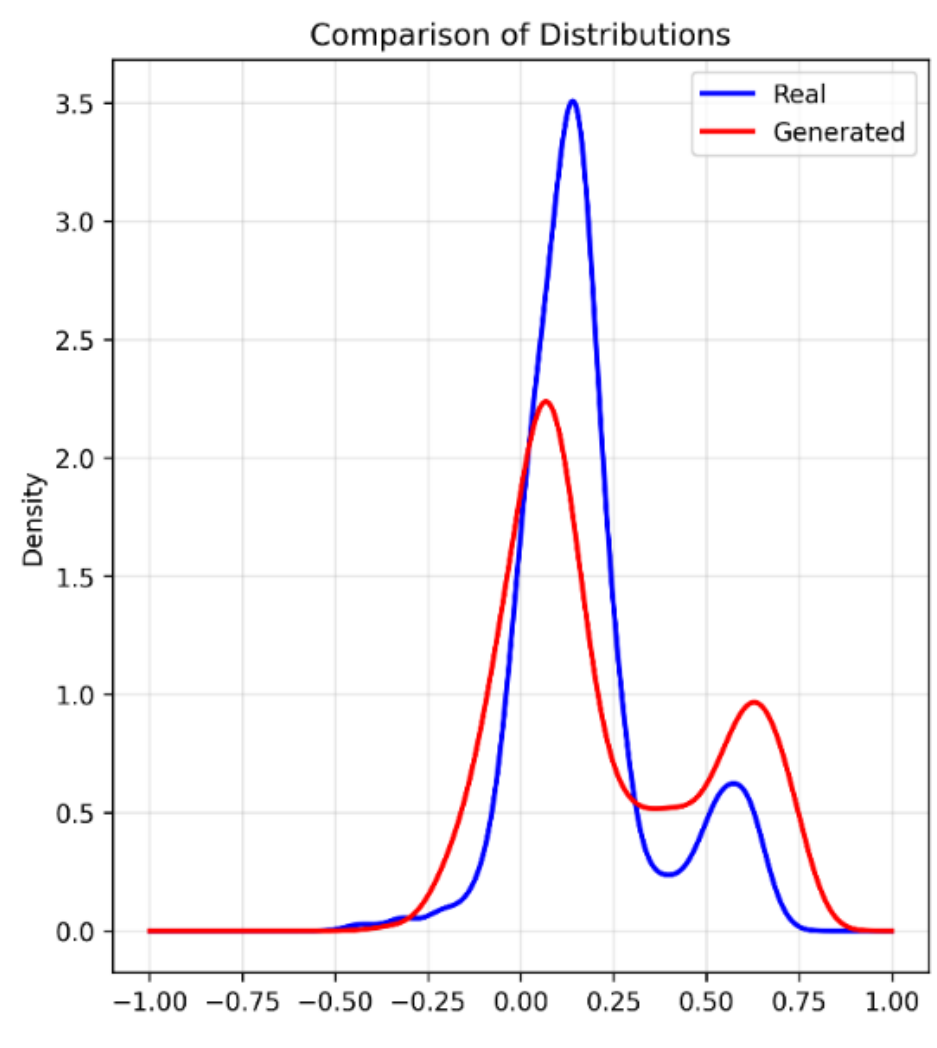

Fig. 24: Comparison between distributions of original and generated data for DCGAN in the continuous case.

***Deep Convolutional Variational Autoencoder (DCVAE)***

The DCVAE model was trained until reach the early stopping criteria. We obtained the FID and FRD after the end of iterations, respectively equal to 330 and 7.65. The training time was only 26 minutes, showing that it is much faster to train the VAE than a GAN. We can see in Fig. 25 the quality of the images reconstructed by the decoder's model, showing images with lower quality than DCGAN, as expected from literature.

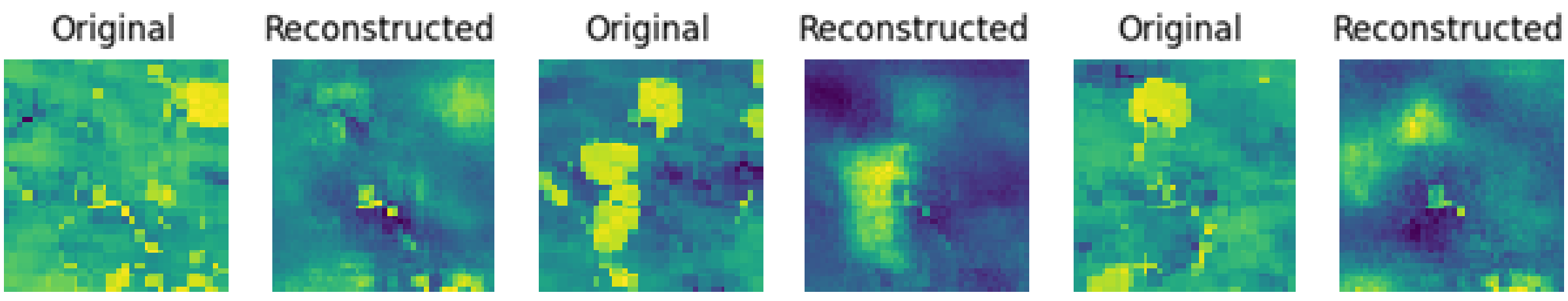

Fig. 25: Three examples of random images from the set along with the reconstructed images by DCVAE in the case study 2.

Fig. 26 show the reconstruction and KL loss curves. The reconstruction loss graph shows a continuous decline in the loss, more intense at the beginning and more slowly thereafter. The KL loss graph shows an initial fast drop followed by a stabilization afterwards with lower errors.

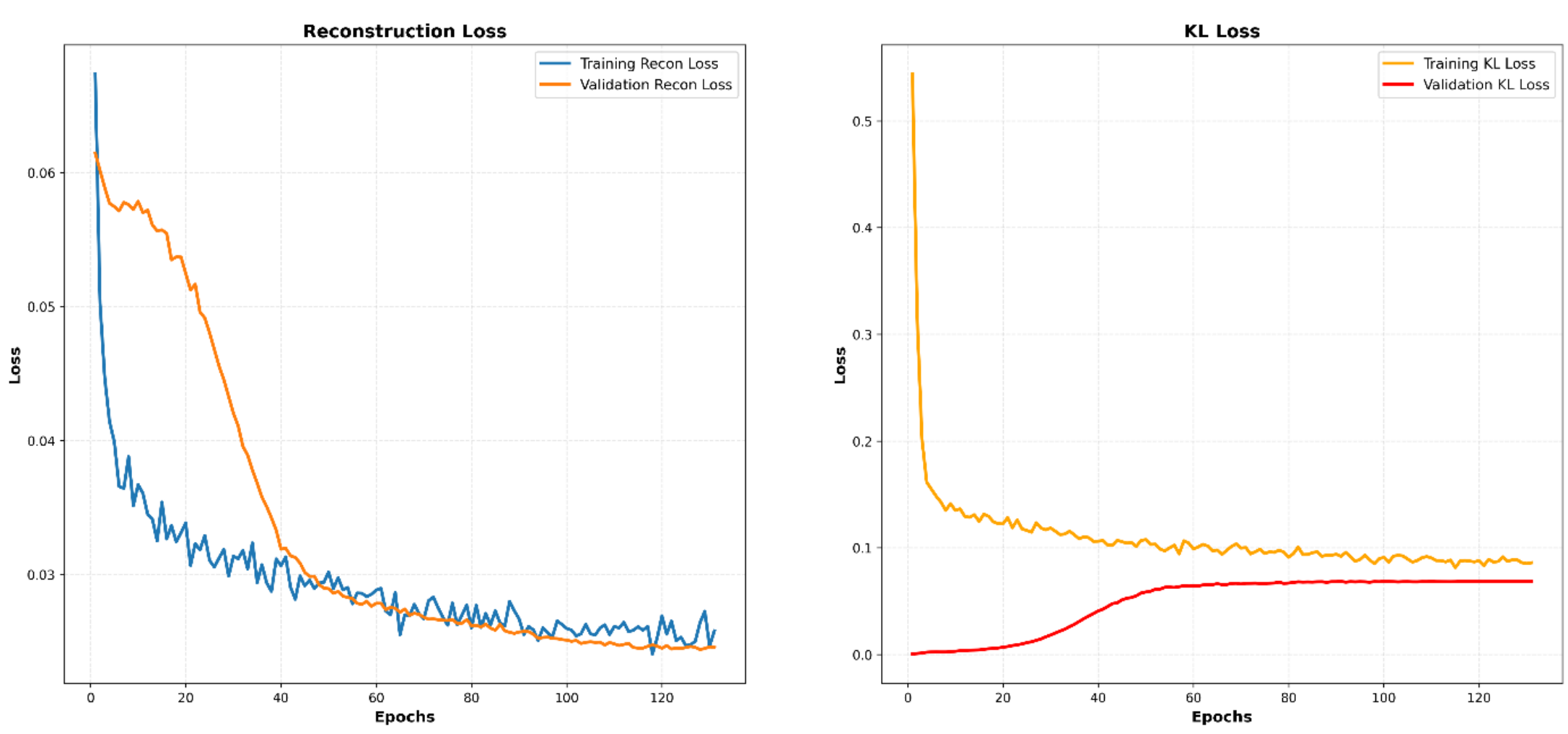

Fig. 26: Reconstruction (left) and Kullback-Leibler loss (right) throughout the training with VAE in the continuous case.

Fig. 27 shows the MSE of the training and validation sets. We can see the continuous decrease throughout the training, stabilizing at the end, showing that 130 epochs were sufficient to train the model.

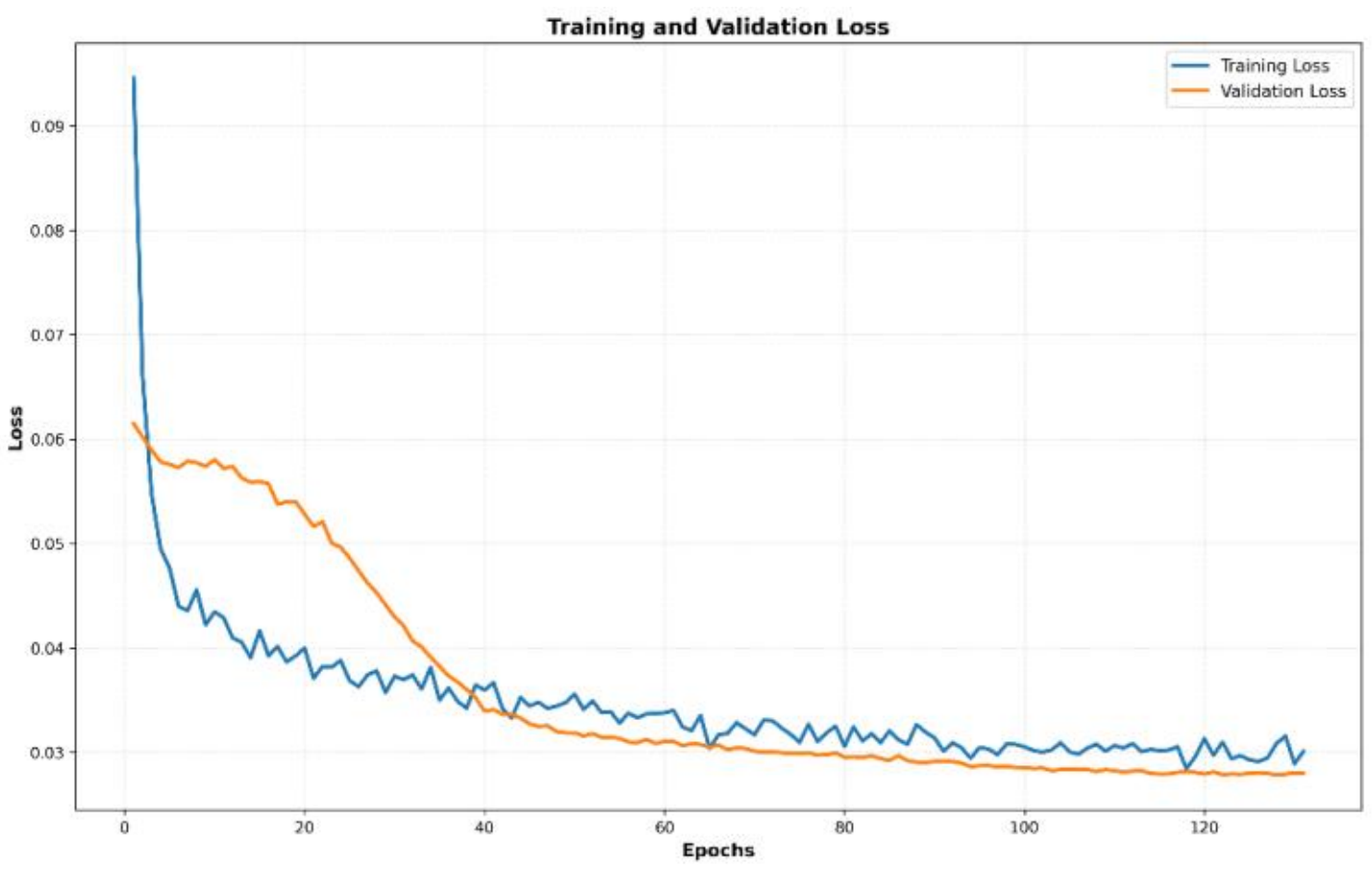

Fig. 27: MSE of training and validation sets versus epochs with VAE in the case study 2.

***Variational Autoencoder Generative Adversarial Network (VAE-GAN)***

The VAE-GAN model was trained until reach the early stopping criteria, reaching the FID of 286 and FRD of 8.32 at the end of iterations. The training time was only 36 minutes, being much faster than GAN also for this case. We can see in Fig. 28 the quality of the images reconstructed by the decoder's model, showing images with higher quality than VAE, and similar or little lower to GAN.

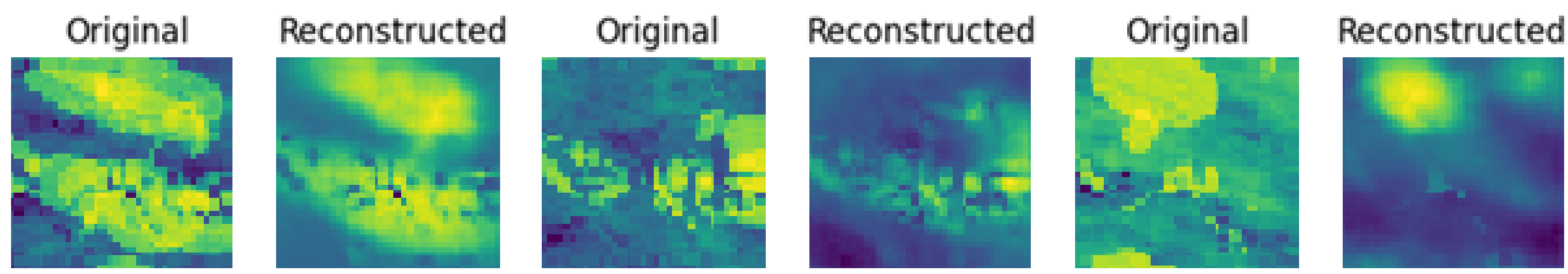

Fig. 28: Three examples of random images from the set along with the reconstructed images by VAE-GAN in the continuous case.

Fig. 29 show the reconstruction and KL loss curves. The reconstruction loss graph shows a continuous decline in the loss, more intense at the beginning and more slowly thereafter. The KL loss graph shows an initial rise destabilization followed by stabilization.

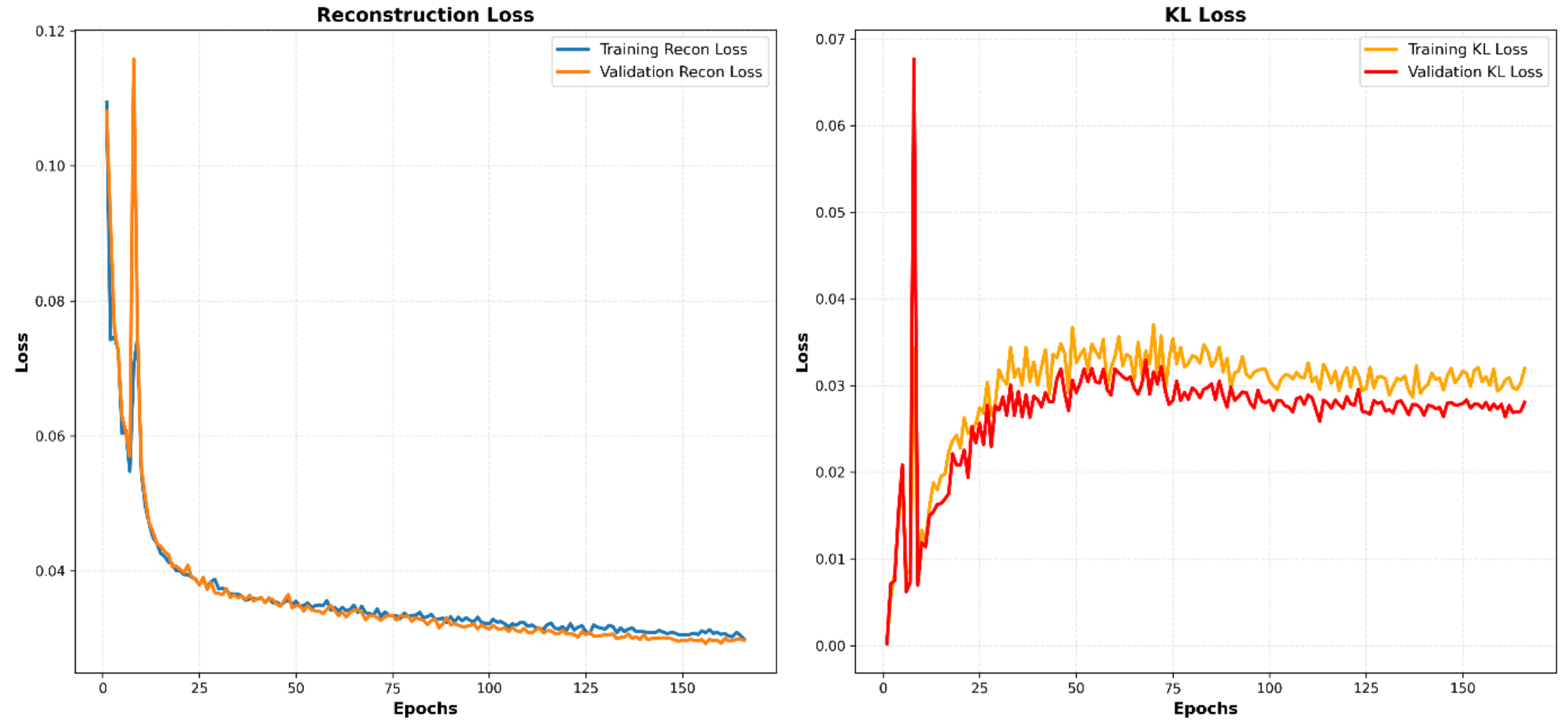

Fig. 29: Graphs of reconstruction (left) and Kullback-Leibler loss (right) throughout the training with VAE-GAN in the case study 2.

Fig. 30 shows the total error curves for the training and the validation sets. This graph is the combination of all losses of this model: reconstruction, KL, generator and discriminator losses. The curves show a stabilization in errors after 75 epochs.

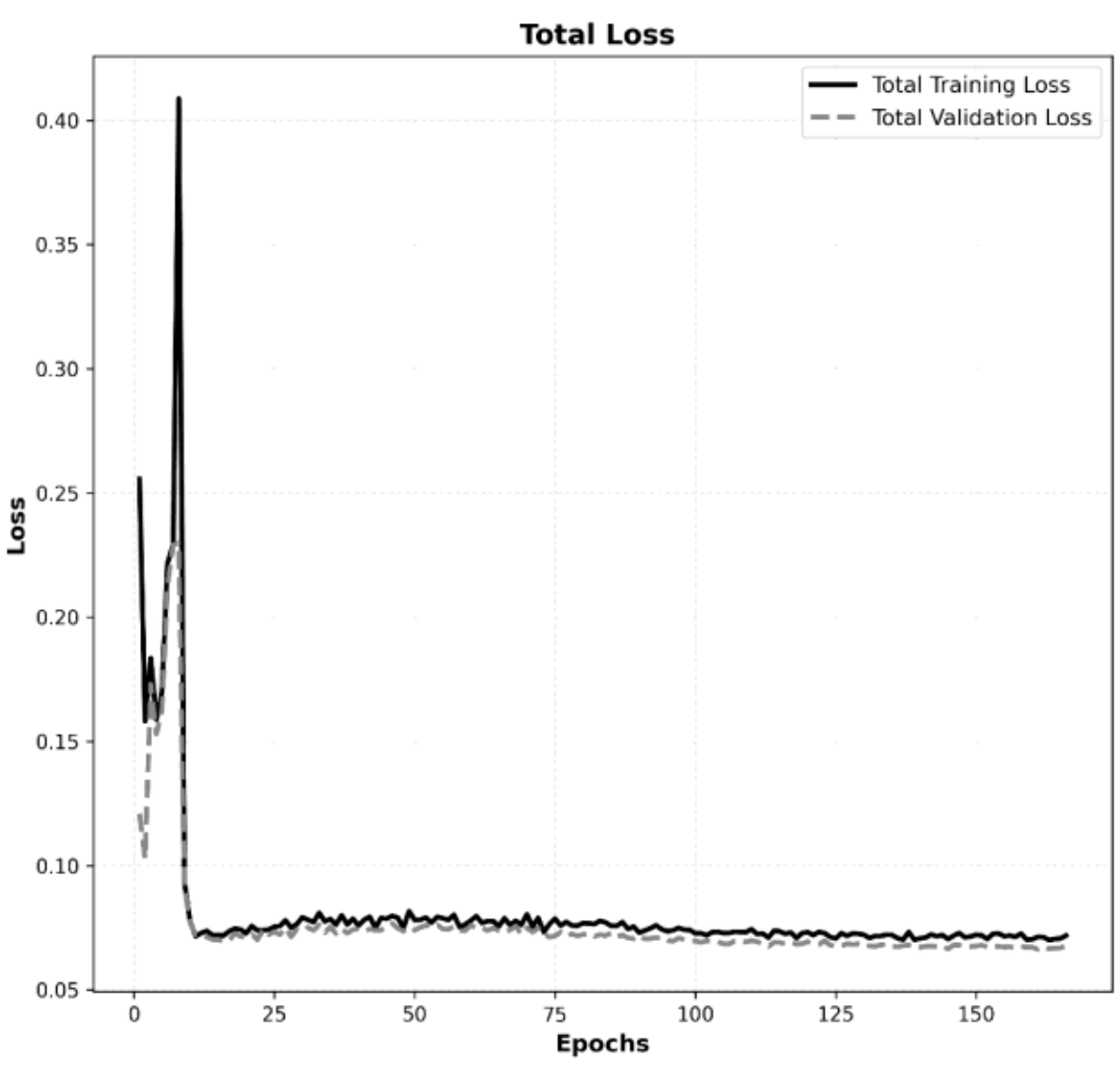

Fig. 30: Total loss versus epochs with VAE-GAN in the continuous case.

In Table 2, we summarize the results of the FID and FRD metrics at the end of training for the three models. We can conclude again that the FRD metric was more appropriate than the FID also for this case study and that the values obtained were low, demonstrating that the training of all models was efficient.

Table 2: Results of FID and FRD in the end of training of models in the continuous case.

| | FID | FRD |
|---|---|---|
| DCGAN | 32.03 | 0.26 |
| DCVAE | 330 | 7.65 |
| VAE-GAN | 286 | 8.32 |

#### *4.2.2. Data Assimilation*

Fig. 31 show the images of the true case, the priori and posteriori mean and standard deviation. We can observe that the posteriori mean presents an image more similar to the true case in the case of VAE and VAE-GAN. We can also observe that the VAE-GAN model can preserve a little more of the ensemble variance than the other models.

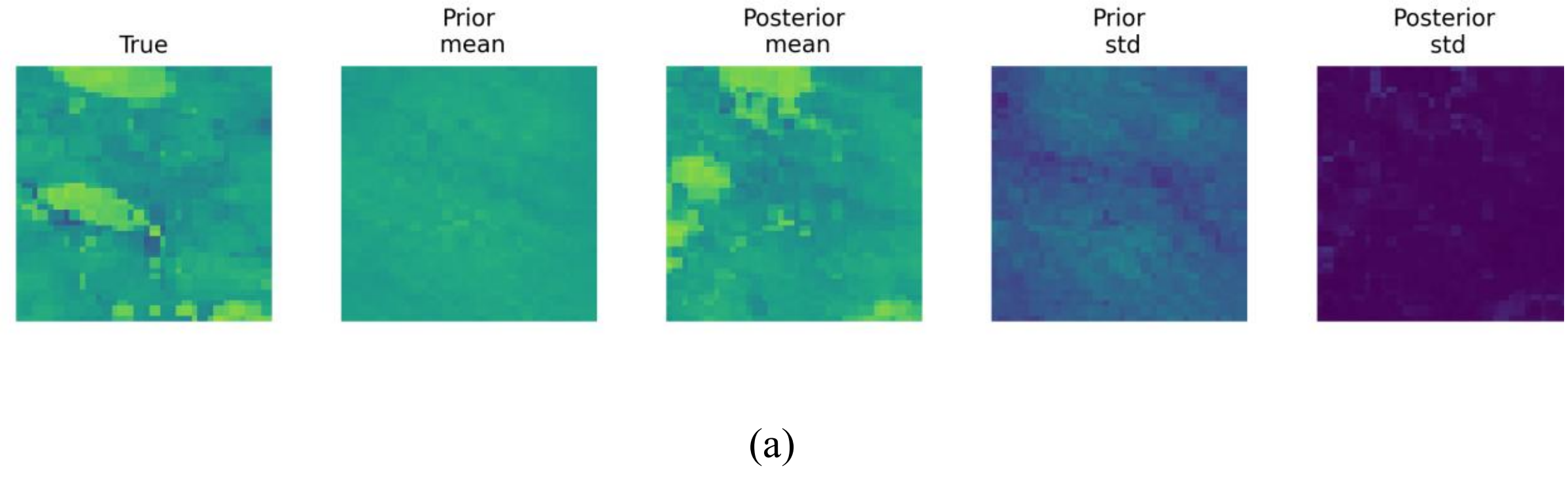

(a)

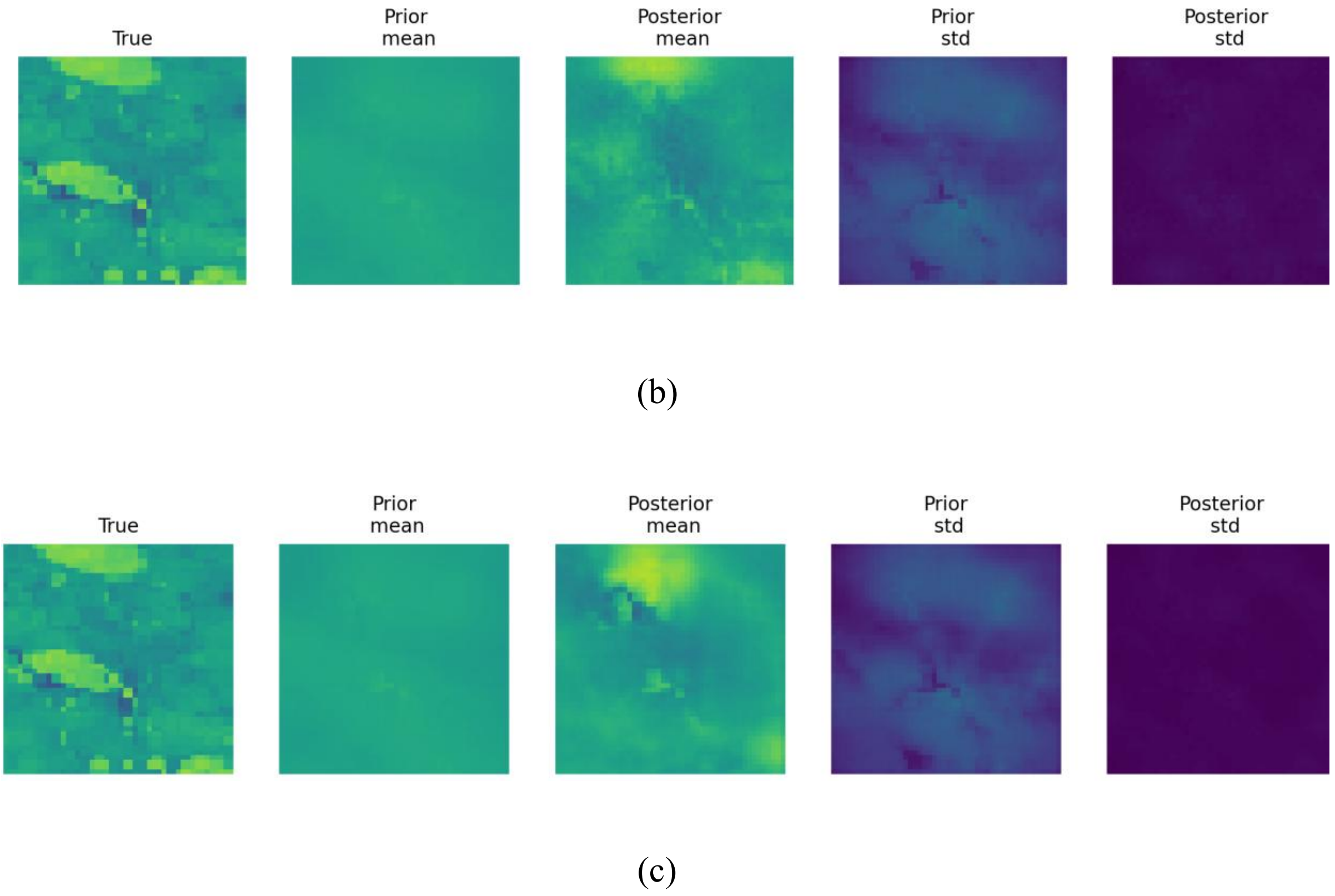

Fig. 31: Images of the true case, the priori and posteriori mean, and the priori and posteriori standard deviation in the case study 2 for: (a) DCGAN, (b) DCVAE and (c) VAE-GAN.

In Fig. 32, we can see the images of the priori and posteriori of member 1, as an example, to illustrate a case before and after assimilation for the second case study. As we can see, the GAN network fails to preserve the features of member 1. In VAE, the network can delineate the model's features, but with low quality. In the VAE-GAN model, the network can delineate the facies contours a little better than the VAE network.

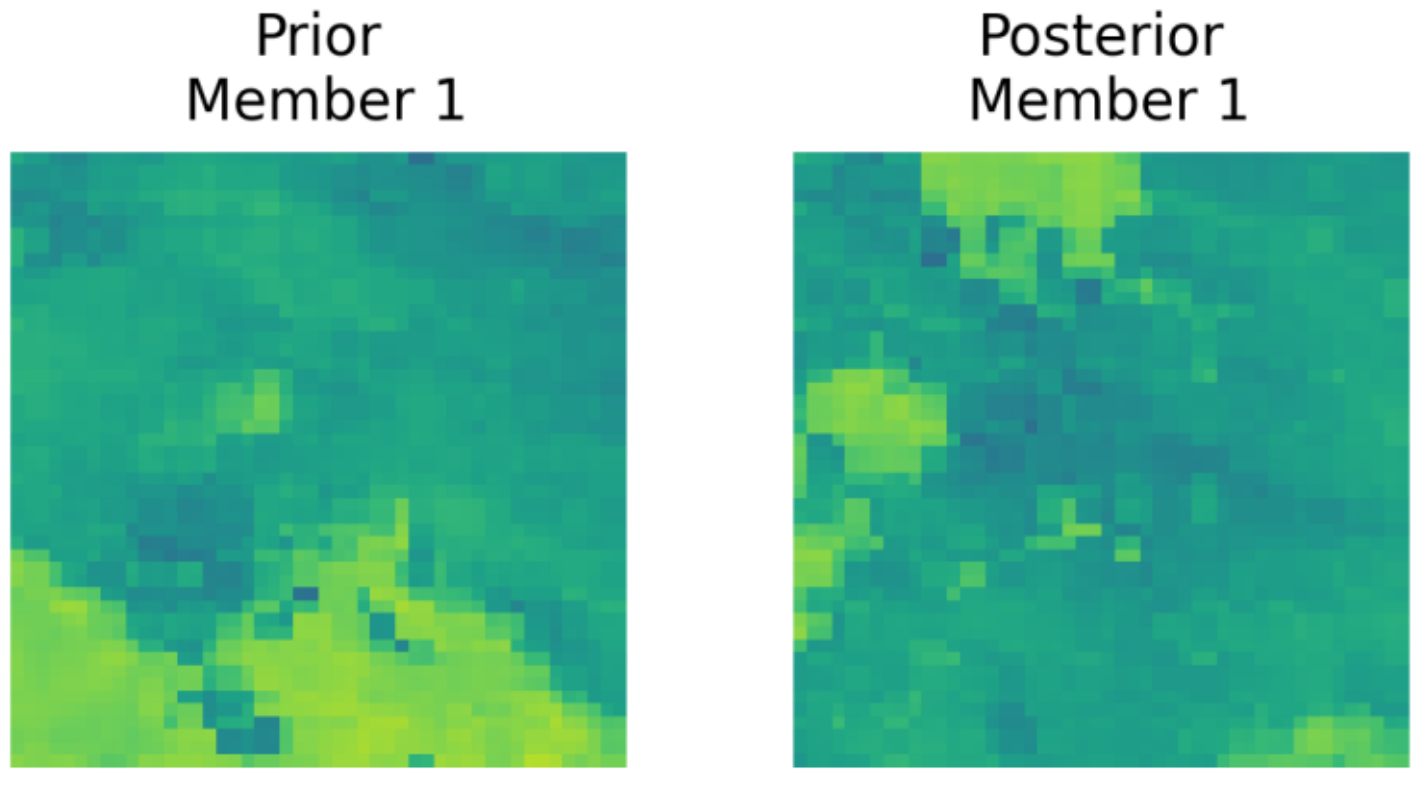

(a)

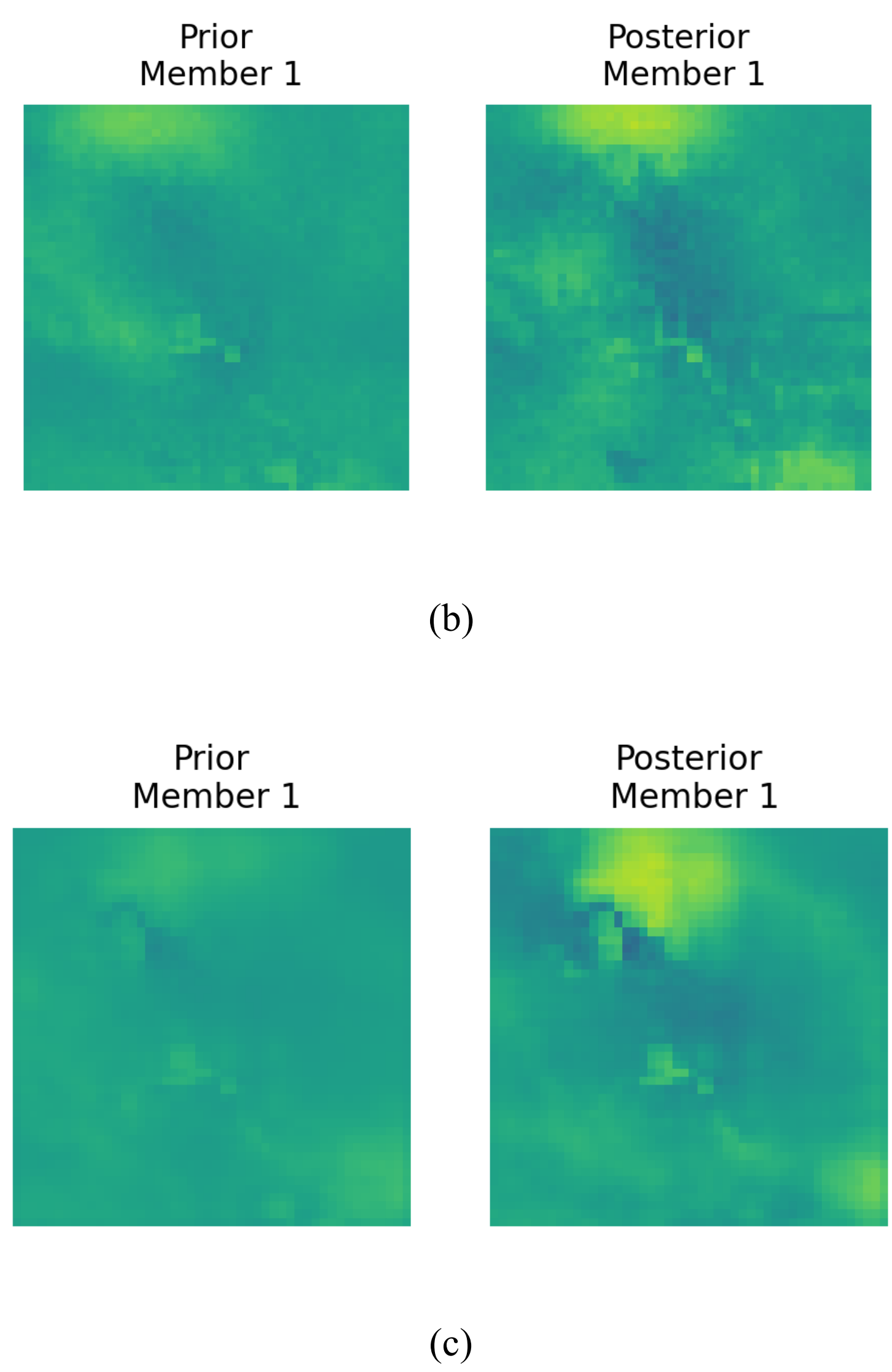

(b)

(c)

Fig. 32: Images of the priori and posteriori of member 1 in the continuous case for: (a) DCGAN, (b) DCVAE and (c) VAE-GAN.

Analyzing the time series of production data in the Fig. 33, it is possible to confirm that no ensemble collapse occurred. Also in this case, we can see a better matching for the VAE, followed by the VAE-GAN model. This can be explained by the ease of the VAE and VAE-GAN models in making the latent vector follow a Gaussian distribution, while we saw more difficulty for GAN model in transforming the original distribution.

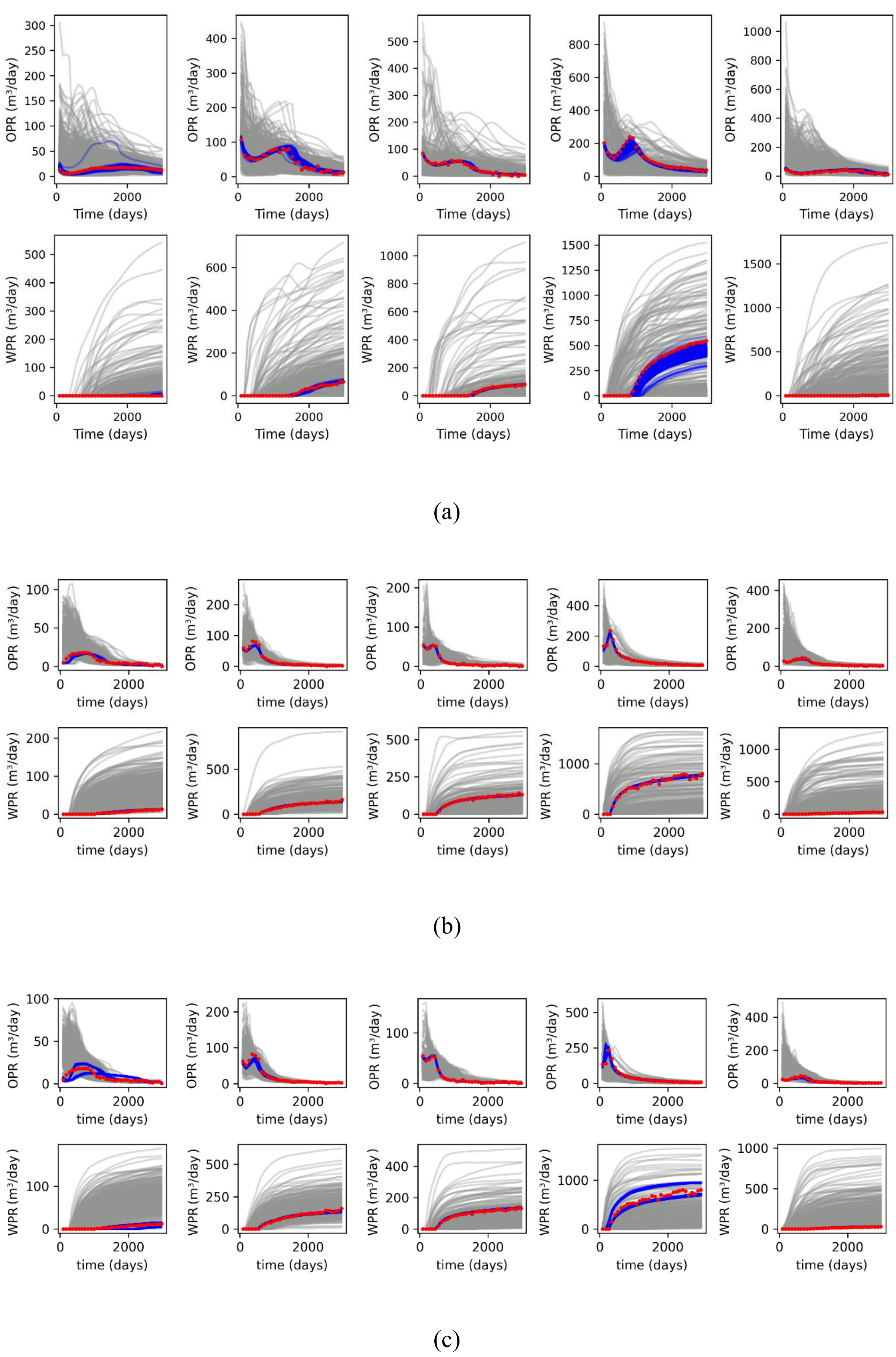

Fig. 33: Time series of production data from the first five producers of oil production rate

(above) and water production rate (below). Here, the gray lines represent the prior ensemble, blue lines represent the posterior ensemble, and the red dots represent the measurements in the case study 2 for: (a) DCGAN, (b) DCVAE and (c) VAE-GAN.

Fig. 34 shows graphs of data mismatching, RMSE, spread as a function of the number of iterations. We can observe that VAE-GAN obtained the best results for DM and spread, while GAN had the best result for RMSE.

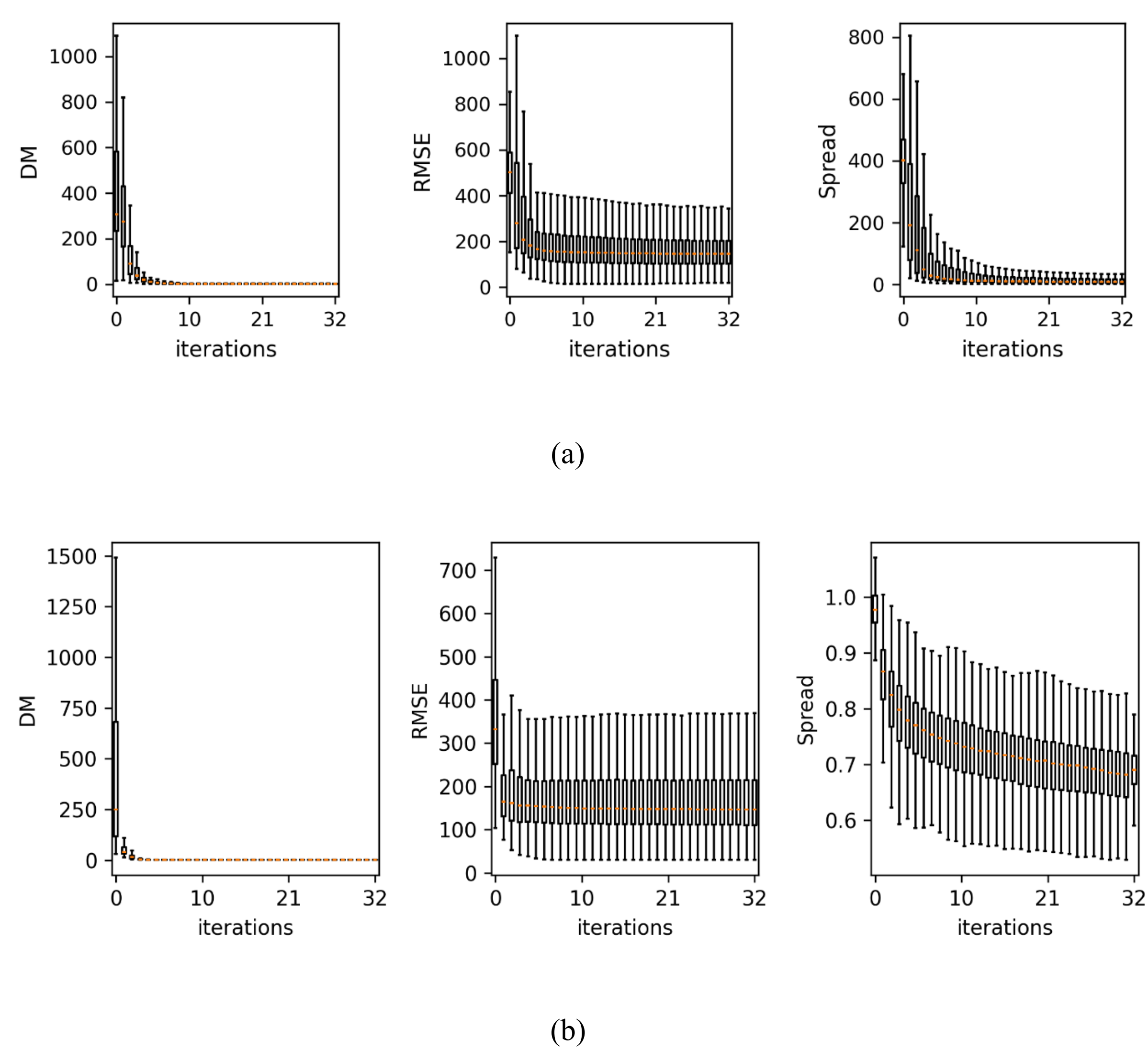

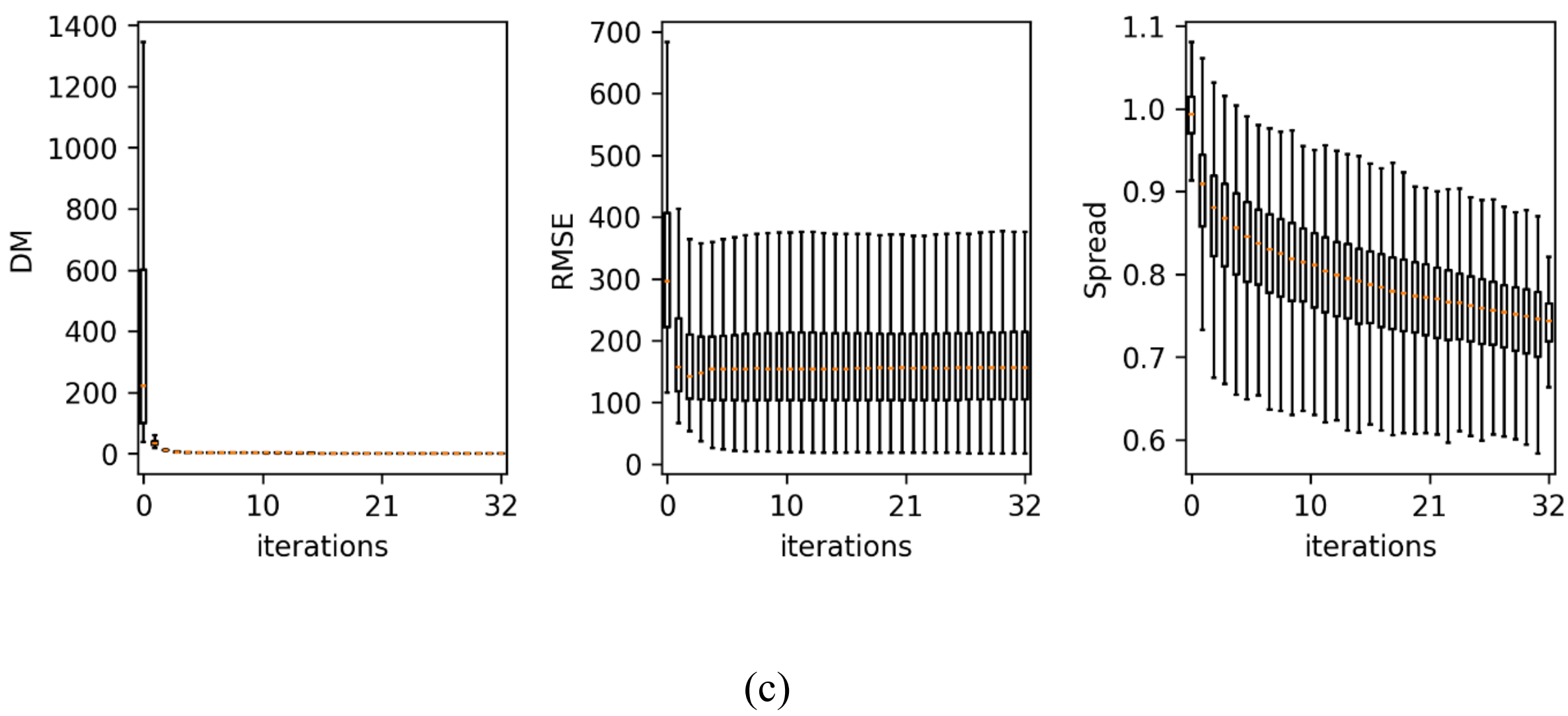

(c)

Fig. 34: Graphs of data mismatching, RMSE and spread as a function of the number of iterations for: (a) DCGAN, (b) DCVAE and (c) VAE-GAN.

### 4.3. Results of Static Geostatistical Metrics

To evaluate the quality of images generated by deep learning models in geological terms, we used main geostatistical statistics metrics, such as: variogram (MSE), connectivity (MSE), histogram KL, PCA correlation and MDS MMD (Maximum Mean Discrepancy). These metrics are defined and available for consultation on GitHub at the following link. In Table 3, we can see the results for static geostatistical metrics. Based on the comparative analysis of static geostatistical metrics, all three generative models demonstrate competent performance in capturing spatial geological structures, with each architecture showcasing specific strengths that merit consideration for different application contexts. DCGAN emerges as the most balanced performer, achieving the lowest Variogram MSE in both discrete (0.0025) and continuous (0.0006) cases, representing superior spatial correlation fidelity—the most critical metric for geological modeling. DCVAE displays notable strength in connectivity preservation for discrete scenarios (0.0045), suggesting particular suitability for binary or categorical geological facies modeling where connected pathways are essential. Meanwhile, VAE-GAN achieves the highest PCA correlation in continuous cases (0.0572), indicating effective capture of principal variation patterns within continuous permeability fields.

Table 3: Results for static geostatistical metrics.

| | Discrete Case | | | Continuous Case | | |
|---|---|---|---|---|---|---|
| | DCGAN | DCVAE | VAE-GAN | DCGAN | DCVAE | VAE-GAN |
| Variogram MSE | 0.0025 | 0.0150 | 0.0187 | 0.0006 | 0.0010 | 0.0008 |
| Connectivity MSE | 0.0079 | 0.0045 | 0.0103 | 0.0005 | 0.0114 | 0.0044 |
| Histogram KL | 0.2658 | 1.6121 | 0.9113 | 0.0162 | 0.7358 | 0.4149 |
| PCA Correlation | 0.0326 | 0.0349 | 0.0333 | 0.0399 | 0.0362 | 0.0572 |
| MDS MMD | 0.1766 | 0.0633 | 0.1369 | 0.0750 | 0.3566 | 0.2025 |

## 5. CONCLUSIONS

We can conclude, as already noticed in the literature, that GAN tends to generate better images and VAE a better match to data. Confirming our initial hypothesis, the VAE-GAN model can generate better images than VAE and better data match than GANs in the same time. In this way, the hybrid model was able to gather high-quality images with an accurate history match in two case studies. The hybrid model can transform a non-Gaussian distribution to a Gaussian one much more easily than GAN, as seen in the results. The hybrid model has a discriminator network that helps the model to find images with better quality and geological realism. The limitations of the model are the difficulty in finding the best configuration for a more complex network and optimizing hyperparameters. This can lead to instabilities during training. Training time can be reduced depending on the case and parameters used, showing feasibility for larger 3D cases. The results of the static geostatistical metrics confirmed the high quality of the geological images generated by the models overall. The results proved to be very promising for application in large-scale reservoirs, for bringing together improvements in matchings along with high-quality images.

**Acknowledgments**

We gratefully acknowledge the support of Escola Politécnica of the University of São Paulo. The authors would also like to thank the LASG (Laboratory of Reservoir Simulation and Management) for supporting this research, CMG (Computer Modelling Group Ltd.) for providing the reservoir simulator licenses used in this study and FAPESP – São Paulo Research Foundation (16/08801-0). We also thank the National Council for Scientific and Technological Development (CNPq) for financial support through grant number 310676/2025-8.

**Conflicts of interest**

The authors declare that they have no known competing financial interests or personal relationships that could have appeared to influence the work reported in this paper.

**Computer Code Availability**

The codes and the datasets generated and/or analyzed during the current study are available in the Github repository: https://github.com/LASG-USP/VAE-GAN

List of Figures
1. Fig. 35: A typical structure of a GAN.
2. Fig. 2: A typical structure of VAE.
3. Fig. 3: A typical structure of VAE-GAN model.
4. Fig. 4: The complete workflow of integration between generative models and ESMDA. In the DCGAN case, the representation of decoder is the generator network.
5. Fig. 5: Random realizations of the categorical training dataset.
6. Fig. 6: The true reservoir model for the first case. The locations of injectors are indicated by triangles, and those of producers by circles.
7. Fig. 7: Random realizations of the continuous training dataset.
8. Fig. 8: The true reservoir model for the second case. The locations of injectors are indicated by triangles, and those of producers by circles.

9. Fig. 9: Images generated by DCGAN after training with 150 000 iterations for case study 1.
10. Fig. 10: Comparison between distributions of (a) original and (b) generated data for DCGAN in the categorical case after training.
11. Fig. 11: Three examples of random images from the set along with the reconstructed images by DCVAE for case study 1.
12. Fig. 12: Graphs of reconstruction (left) and Kullback-Leibler loss (right) throughout the training with DCVAE for case study 1.
13. Fig. 13: MSE of training and validation versus epochs with DCVAE across training iterations.
14. Fig. 14: Comparison between distributions of original data and latent space for VAE in the case study 1.
15. Fig. 15: Three examples of random images from the set along with the reconstructed images by VAE-GAN for case study 1.
16. Fig. 16: Reconstruction and Kullback-Leibler losses throughout the training with VAE-GAN in the categorical case.
17. Fig. 17: Total loss versus epochs with VAE-GAN in the categorical case.
18. Fig. 18: Comparison between distributions of original data and latent space for VAE-GAN in the categorical case.
19. Fig. 19: Images of the true case, the priori and posteriori mean, and the priori and posteriori standard deviation in the categorical case for: (a) DCGAN, (b) DCVAE and (c) VAE-GAN.
20. Fig. 20: Images of the priori and posteriori of member 1 in the categorical case for: (a) DCGAN, (b) DCVAE and (c) VAE-GAN.
21. Fig. 21: Time series of production data from the first five producers of oil production rate (above) and water production rate (below). Here, the gray lines represent the prior ensemble, blue lines represent the posterior ensemble, and the red dots represent the measurements in the case study 2 for: (a) DCGAN, (b) DCVAE and (c) VAE-GAN.

34. Fig. 34: Graphs of data mismatching, RMSE and spread as a function of the number of iterations for: (a) DCGAN, (b) DCVAE and (c) VAE-GAN.

## Appendix: Architecture of the networks

Table 4: Configuration of generator of the DCGAN model for both cases.

| **Layer** | **Output / shape** | **Obs.** |
|---|---|---|
| **Generator** | | |
| Input | Shape = (100) | Input latent vector |
| Dense | (12288) | Projection to initial size |
| Reshape | (3, 3, 512) | Reshape output dimension |
| Batch Normalization | (3, 3, 512) | Regularization |
| ReLU | (3, 3, 512) | Activation |
| Convolution 2D | (3, 3, 512) | Kernels=512, size=(3,3) |
| Batch Normalization | (3, 3, 512) | Regularization |
| ReLU | (3, 3, 512) | Activation |
| Convolution 2D | (6, 6, 256) | Kernels=256, size=(3,3) |
| Batch Normalization | (6, 6, 256) | Regularization |
| ReLU | (6, 6, 256) | Activation |
| Convolution 2D | (12, 12, 128) | Kernels=128, size=(3,3) |
| Batch Normalization | (12, 12, 128) | Regularization |
| ReLU | (12, 12, 128) | Activation |
| Convolution 2D | (24, 24, 64) | Kernels=64, size=(3,3) |
| Batch Normalization | (24, 24, 64) | Regularization |
| ReLU | (24, 24, 64) | Activation |
| Convolution 2D | (48, 48, 32) | Kernels=32, size=(3,3) |
| ReLU | (48, 48, 32) | Activation |
| Convolution 2D | (48, 48, 16) | Kernels=16, size=(3,3) |
| ReLU | (48, 48, 32) | Activation |
| Convolution 2D | (48, 48, 1) | Kernels=1, size=(1,1) |
| Tanh | (48, 48, 1) | Activation |

| Resize | (48, 48, 1) | Ensure output shape |
|---|---|---|

Table 5: Configuration of discriminator of the DCGAN model for both cases.

| **Layer** | **Output / shape** | **Obs.** |
|---|---|---|
| **Discriminator** | | |
| Input | Shape = (48, 48, 1) | Input |
| Convolution 2D | (48, 48, 64) | Kernels=64, size=(1,1) |
| Leaky ReLU | (48, 48, 64) | α=0.2 |
| Convolution 2D | (48, 48, 128) | Kernels=128, size=(3,3) |
| Leaky ReLU | (48, 48, 128) | α=0.2 |
| Average Pooling 2D | (24, 24, 128) | Pool size=(2,2), stride=(2,2) |
| Convolution 2D | (24, 24, 256) | Kernels=256, size=(3,3) |
| Leaky ReLU | (24, 24, 256) | α=0.2 |
| Average Pooling 2D | (12, 12, 256) | Pool size=(2,2), stride=(2,2) |
| Convolution 2D | (12, 12, 512) | Kernels=512, size=(3,3) |
| Leaky ReLU | (12, 12, 512) | α=0.2 |
| Average Pooling 2D | (6, 6, 512) | Pool size=(2,2), stride=(2,2) |
| Convolution 2D | (6, 6, 512) | Kernels=512, size=(3,3) |
| Leaky ReLU | (6, 6, 512) | α=0.2 |
| Average Pooling 2D | (3, 3, 512) | Pool size=(2,2), stride=(2,2) |
| Convolution 2D | (3, 3, 1024) | Kernels=1024, size=(3,3) |
| Leaky ReLU | (3, 3, 1024) | α=0.2 |
| Global Average Pooling 2D | (1024) | Spatial pooling |
| Fully-connected | (1) | Output score, activation=linear |

Table 6: Configuration of encoder of the DCVAE model for first case.

| Layer | Output / shape | Obs. |
| --- | --- | --- |
| **Encoder** | | |
| Input | Shape = (51, 51, 2) | Input data |
| Convolution 2D | (26, 26, 32) | Kernels=32, size=(3,3), stride=2, padding="same" |
| LeakyReLU | (26, 26, 32) | Activation, α=0.2 |
| Batch Normalization | (26, 26, 32) | Regularization |
| Convolution 2D | (13, 13, 64) | Kernels=64, size=(3,3), stride=2, padding="same" |
| LeakyReLU | (13, 13, 64) | Activation, α=0.2 |
| Batch Normalization | (13, 13, 64) | Regularization |
| Convolution 2D | (7, 7, 128) | Kernels=128, size=(3,3), stride=2, padding="same" |
| LeakyReLU | (7, 7, 128) | Activation, α=0.2 |
| Batch Normalization | (7, 7, 128) | Regularization |
| Convolution 2D | (4, 4, 256) | Kernels=256, size=(3,3), stride=2, padding="same" |
| LeakyReLU | (4, 4, 256) | Activation, α=0.2 |
| Batch Normalization | (4, 4, 256) | Regularization |
| Dropout | (4, 4, 256) | Rate=0.4, regularization |
| Flatten | (4096) | Setup for the fully-connected layer |
| Fully-connected | (256) | ---- |
| LeakyReLU | (256) | Activation, α=0.2 |
| Dropout | (256) | Rate=0.3, regularization |
| Fully-connected | Neurons = 512 | Mean of the VAE |
| Fully-connected | Neurons = 512 | Log-variance of the VAE |
| Fully-connected | Neurons = 512 | $z = \mu + \varepsilon \times \exp(\sigma^2/2)$ |

Table 7: Configuration of decoder of the DCVAE model for first case.

| Layer | Output / shape | Obs. |
|---|---|---|
| **Decoder** | | |
| Input | Shape = 512 | Input latent vector |
| Fully-connected | (9216) | 6×6×256 = 9216 |
| LeakyReLU | (9216) | Activation, α=0.2 |
| Dropout | (9216) | Rate=0.2, regularization |
| Reshape | (6, 6, 256) | Reshape output dimension |
| Convolution 2D Transpose | (12, 12, 128) | Kernels=128, size=(3,3), stride=2, padding="same" |
| LeakyReLU | (12, 12, 128) | Activation, α=0.2 |
| Batch Normalization | (12, 12, 128) | Regularization |
| Convolution 2D Transpose | (24, 24, 64) | Kernels=64, size=(3,3), stride=2, padding="same" |
| LeakyReLU | (24, 24, 64) | Activation, α=0.2 |
| Batch Normalization | (24, 24, 64) | Regularization |
| Convolution 2D Transpose | (48, 48, 32) | Kernels=32, size=(3,3), stride=2, padding="same" |
| LeakyReLU | (48, 48, 32) | Activation, α=0.2 |
| Batch Normalization | (48, 48, 32) | Regularization |
| Convolution 2D | (48, 48, 2) | Kernels=2, size=(3,3), activation="tanh", padding="same" |

Table 8: Differences between decoder architectures in both cases.

| Feature | First case | Second case | Change |
|---|---|---|---|
| Dense Layer | 6×6×512 = 18432 | 6×6×256 = 9216 | Reduced 50% |
| Dropout | Not present | Rate=0.2 | Added |

| Feature | First case | Second case | Change |
|---|---|---|---|
| Conv Layers | 3 simple layers | 3 layers with BN | Improved regularization |
| Extra Layers | Had refinement layers | Removed | Simplified |

Table 9: Configuration of encoder of the VAE-GAN model for first case.

| **Layer** | **Output / shape** | **Obs.** |
|---|---|---|
| **Encoder** | | |
| Input | Shape = (51, 51, 2) | Input data |
| Convolution 2D | (26, 26, 58) | Kernels=58, size=(5,5), stride=2, padding="same" |
| LeakyReLU | (26, 26, 58) | Activation, α=0.2 |
| Batch Normalization | (26, 26, 58) | Regularization |
| Convolution 2D | (13, 13, 116) | Kernels=116, size=(5,5), stride=2, padding="same" |
| LeakyReLU | (13, 13, 116) | Activation, α=0.2 |
| Batch Normalization | (13, 13, 116) | Regularization |
| Convolution 2D | (7, 7, 230) | Kernels=230, size=(5,5), stride=2, padding="same" |
| LeakyReLU | (7, 7, 230) | Activation, α=0.2 |
| Batch Normalization | (7, 7, 230) | Regularization |
| Flatten | (7×7×230) = 11270 | Setup for the fully-connected layer |
| Fully-connected | Neurons = 922, activation = Leaky ReLU | ---- |
| Dropout | 30% | Strategy to avoid overfitting |
| Fully-connected | Neurons = 512 | Mean of the VAE-GAN |
| Fully-connected | Neurons = 512 | Log-variance of the VAE-GAN |
| Z (lambda) | Neurons = 512 | Latent layer |

Table 10: Configuration of decoder of the VAE-GAN model for first case.

| Layer | Output / shape | Obs. |
|---|---|---|
| **Decoder** | | |
| Input | Shape = 512 | Input latent vector |
| Fully-connected | (6×6×230) = 8280 | ---- |
| LeakyReLU | (8280) | Activation, α=0.2 |
| Reshape | Output size = (6, 6, 230) | Reshape output dimension |
| Convolution 2D Transpose | (12, 12, 230) | Kernels=230, size=(5,5), stride=2, padding="same" |
| LeakyReLU | (12, 12, 230) | Activation, α=0.2 |
| Batch Normalization | (12, 12, 230) | Regularization |
| Convolution 2D Transpose | (24, 24, 116) | Kernels=116, size=(5,5), stride=2, padding="same" |
| LeakyReLU | (24, 24, 116) | Activation, α=0.2 |
| Batch Normalization | (24, 24, 116) | Regularization |
| Convolution 2D Transpose | (48, 48, 58) | Kernels=58, size=(5,5), stride=2, padding="same" |
| LeakyReLU | (48, 48, 58) | Activation, α=0.2 |
| Batch Normalization | (48, 48, 58) | Regularization |
| Convolution 2D Transpose | (48, 48, 2) | Kernels=2, size=(5,5), activation="tanh", padding="same" |

Table 11: Configuration of discriminator of the VAE-GAN model for first case.

| Layer | Output / shape | Obs. |
|---|---|---|
| **Discriminator** | | |
| Input | Shape = (51, 51, 2) | Input image |
| Gaussian Noise | (51, 51, 2) | Noise std=0.1, prevents discriminator overconfidence |
| SpectralNorm(Conv2D) | (26, 26, 58) | Kernels=58, size=(5,5), stride=2, padding="same" |
| LeakyReLU | (26, 26, 58) | Activation, α=0.2 |
| Dropout | (26, 26, 58) | Rate=0.3, regularization |
| SpectralNorm(Conv2D) | (13, 13, 116) | Kernels=116, size=(5,5), stride=2, padding="same" |
| LeakyReLU | (13, 13, 116) | Activation, α=0.2 |
| Dropout | (13, 13, 116) | Rate=0.3, regularization |
| SpectralNorm(Conv2D) | (7, 7, 230) | Kernels=230, size=(5,5), stride=2, padding="same" |
| LeakyReLU | (7, 7, 230) | Activation, α=0.2 |
| Dropout | (7, 7, 230) | Rate=0.3, regularization |
| Flatten | (11270) | Setup for the fully-connected layer |
| Fully-connected | (1) | Output score |